\pdfoutput=1

\documentclass[11pt]{article}

\usepackage{emnlp2021}

\usepackage{times}
\usepackage{latexsym}
\usepackage{amsmath}
\usepackage[T1]{fontenc}

\usepackage[utf8]{inputenc}

\usepackage{multirow}
\usepackage{amssymb}
\usepackage{stackengine}
\usepackage{graphicx}
\usepackage{microtype}

\newcommand{\myparagraph}[1]{\paragraph{#1}}
\title{{I}diosyncratic but not Arbitrary: {L}earning Idiolects in Online Registers Reveals Distinctive yet Consistent Individual Styles}

\author{Jian Zhu \\
  Department of Linguistics \\
  University of Michigan \\
  \texttt{lingjzhu@umich.edu} \\\And
  David Jurgens \\
  School of Information \\
  University of Michigan \\
  \texttt{jurgens@umich.edu} \\}

\begin{document}
\maketitle
\begin{abstract}
An individual's variation in writing style is often a function of both social and personal attributes. While structured social variation has been extensively studied, e.g., gender based variation, far less is known about how to characterize individual styles due to their idiosyncratic nature. We introduce a new approach to studying idiolects through a massive cross-author comparison to identify and encode stylistic features. The neural model achieves strong performance at authorship identification on short texts and through an analogy-based probing task, showing that the learned representations exhibit surprising regularities that encode qualitative and quantitative shifts of idiolectal styles. Through text perturbation, we quantify the relative contributions of different linguistic elements to idiolectal variation. Furthermore, we provide a description of idiolects through measuring inter- and intra-author variation, showing that variation in idiolects is often distinctive yet consistent. 
\end{abstract}

\section{Introduction}
Linguistic identities  manifest through ubiquitous language variation. The notion that language functions as stylistic resources for the construction and performance of social identity rests upon two theoretical constructs: sociolect and idiolect \cite{grant2018resources}. The term `sociolect' refers to the socially structured variation at the group level, whereas `idiolect' denotes language variation 
associated with individuals \cite{wardhaugh2011introduction,grant2018resources}.  Variationist sociolinguistics emphasizes the systematic variation of sociolect such as gender, ethnicity, and socioeconomic stratification \cite{labov1972sociolinguistic}. While  a central concept in sociolinguistics, idiolect has received far more research attention in forensic linguistics \cite{wright2018idiolect,grant2018resources}. 

Although idiolects have played a central role in stylometry and forensic linguistics, which seek to quantify and characterize individual textual features to separate authors \cite{grant2012txt,coulthard2016introduction,neal2017surveying}, the theory of idiolect remains comparatively underdeveloped \cite{grant2018resources}. An in-depth understanding of the nature and the variation of idiolect not only sheds light on the theoretical discussion of language variation but also aids practical and forensic applications of linguistic science.

Here, we characterize the idiolectal variation of linguistic styles through a computational analysis of large-scale short online texts. Specifically, we ask the following questions: 1) to what extent can we extract distinct styles from short texts, even for unseen authors; 2) what are the core stylistic dimensions along which individuals vary; and 3) to what extent are idiolects consistent and distinctive?

\begin{figure*}[tbh]
    \centering
    \includegraphics[width=0.49\linewidth]{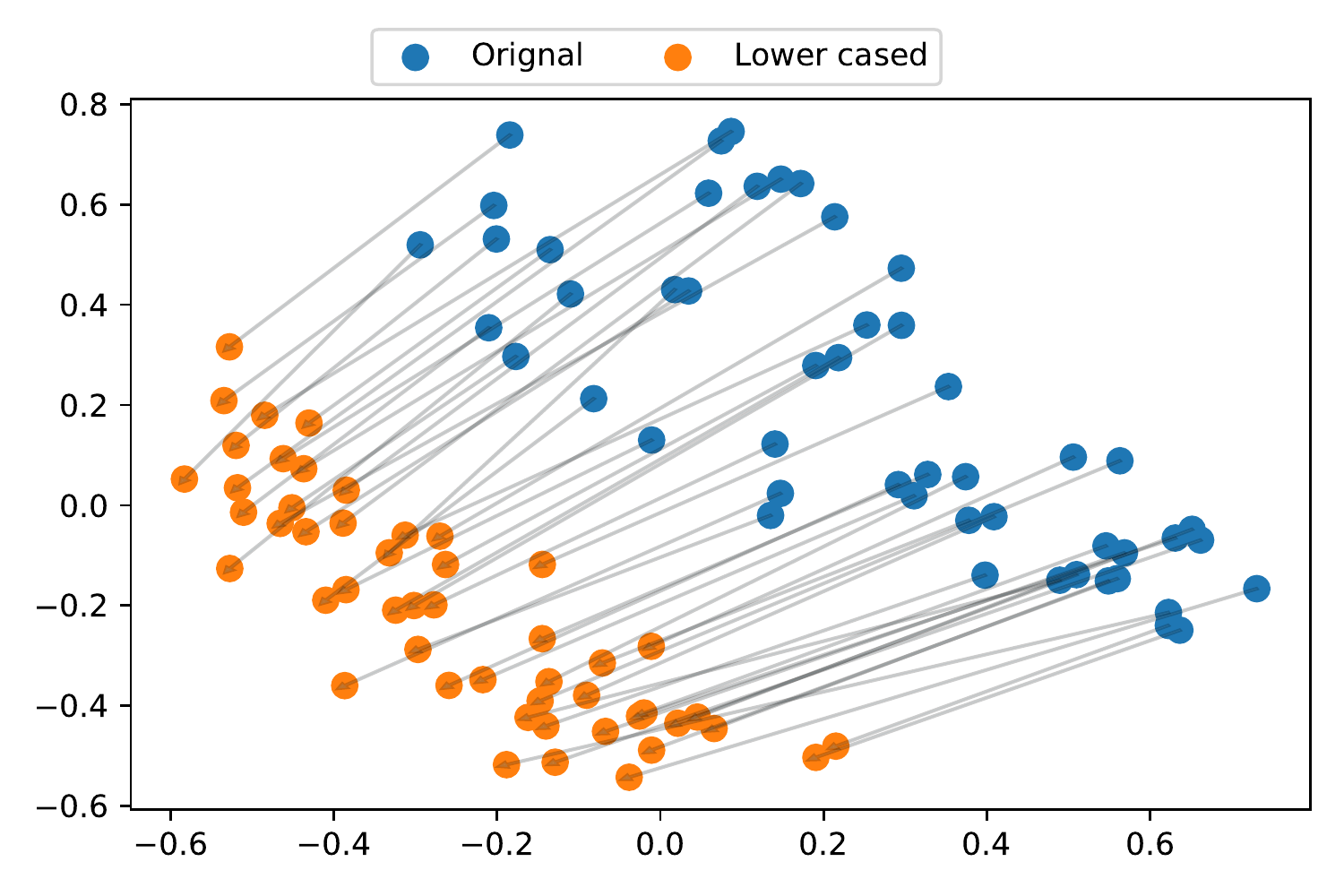}
    \includegraphics[width=0.49\linewidth]{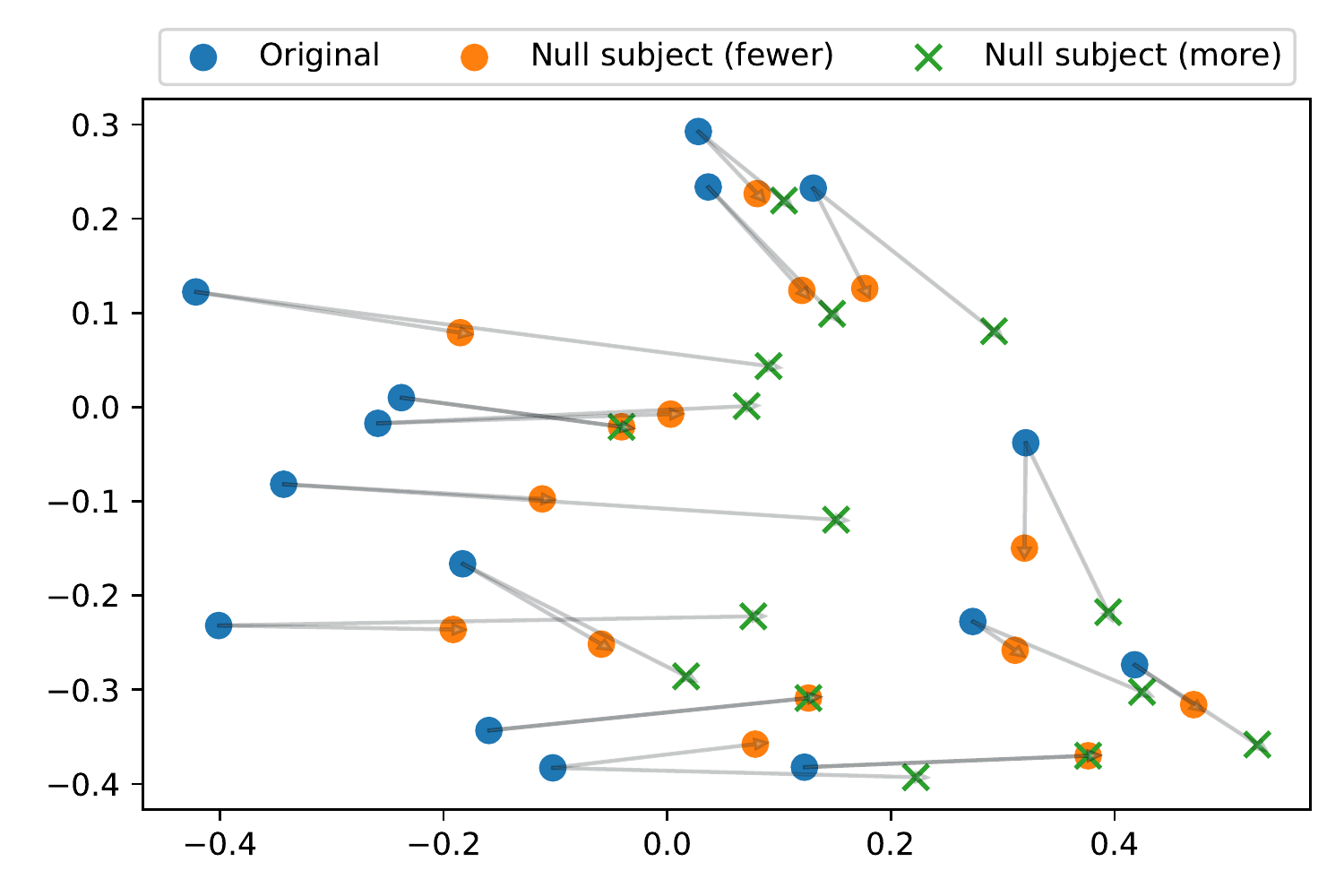}
    \caption{Compositionality in stylistic embeddings encoded by SRoBERTa, projected into the first two principal components. {\bf [Left]} Lower-casing all letters shifts all original texts in the same direction (blue dots $\rightarrow$ orange dots; e.g., ``I love Cantonese BBQ!''$\rightarrow$``i love cantonese bbq!''). {\bf [Right]} The magnitude of movement in one direction is proportionate to the number of null-subject sentences in texts (blue dots $\rightarrow$ orange dots $\rightarrow$ green crosses; e.g., ``I went out. I bought  durians.''$\rightarrow$``Went out. I bought  durians.''$\rightarrow$``Went out. Bought durians''.).}
    \label{fig:composition}
\end{figure*}

By using deep metric learning, we show that idiolect is, in fact, systematic and can be quantified separately from sociolect. And we introduce a new set of probing tasks for testing the relative contributions of different linguistic variations to idiolectal styles. Secondly, we found that the learned representations for idiolect also encode some stylistic dimensions with surprising regularity (see Figure~\ref{fig:composition}), analogous to linguistic regularity found in the word embeddings.
Thirdly, with our proposed metrics for style distinctiveness and consistency, we show that individuals vary considerably in their internal consistency and distinctiveness in idiolect, which has implications for the limits of authorship recognition and the practice of forensic linguistics. For replication, we make our code available at \url{https://github.com/lingjzhu/idiolect}.

\section{Idiolectal variation}
\myparagraph{Theoretical questions}
``Idiolect" remains a fundamental yet elusive construct in sociolinguistics \cite{wright2018idiolect}. The term has been abstractly defined as the totality of the possible utterances one could say \cite{bloch1948set,turell2010use,wright2018idiolect}. Idiolect as a combination of one's cognitive capacity and sociolinguistic experiences \cite{grant2018resources} raises many interesting linguistic questions. First, how are idiolects composed? Forensic studies often focus on a few linguistic features as capturing a person's idiolect \cite{coulthard2004author,barlow2013individual,wright2013stylistic} but few have explicitly offered an explanation as to why particular word sequences were useful or not \cite{wright2017using}. Frameworks for analyzing individual variations have been proposed \cite{grant2012txt,grant2020language} but contributions to idiolectal variation at different linguistic levels are seldom explicitly measured. 

Sociolinguists have pursued the relationship between idiolect and sociolect. However, the perceived idiosyncrasies of idiolect often render it second to sociolect as an object of study \cite{labov1989exact}. Multiple scholars have suggested that idiolects are the building blocks of various sociolects \cite{eckert2012three,barlow2013individual,wright2018idiolect}.
While some studies have probed the relations between the language of individuals and the group \cite{johnstone1997self,schilling1998investigating,meyerhoff2007persistence}, it remains less clear as to what extent idiolect is composed of sociolects or has unique elements of their own \cite[see][for a review]{barlow2013individual}. We test this hypothesize relationship later in Appendix~\ref{app:sociolect} and do find preliminary quantitative evidence that idiolect representations some information of sociolects.

The other theoretical question relevant to cognitive science as well as forensic linguistics is to what extent an individual's idiolect is {\it distinctive} and {\it consistent} against a background population \cite{grant2012txt,grant2018resources}.
Studies on forensic linguistics \cite{johnson2017identifying,wright2013stylistic,wright2017using} provide confirmatory answers to both questions---yet these studies only focus on a specific set of features for a small group of authors. At the other extreme, the high performance of applying machine learning on authorship verification and attribution \cite{kestemont:2018,kestemont:2019,kestemont2020overview} stems from placing more emphasis on separating authors (distinctiveness) than consistency. As an empirical investigation of these two concepts in a relatively large population remains to be conducted. Here we aim to quantify these two linguistic constructs in large-scale textual datasets.

\myparagraph{Stylometry and stylistic similarities}
Traditional stylometry often relies on painstaking manual analysis of texts for a closed set of authors \cite{holmes1998evolution}. Surface linguistic features, especially function words or character n-grams, have been found to be effective in authorship analysis \cite{kestemont-2014-function,neal2017surveying}. Despite the overwhelming success of deep learning, traditional features are still effective in authorship analysis \cite{kestemont:2018,kestemont:2019,kestemont2020overview}.
Yet the wide application of machine learning and deep learning in recent years has greatly advanced the state-of-the-art performance in authorship verification \cite{boenninghoff2019explainable,boenninghoff2019similarity,weerasinghe:2020}. 
Recent PAN Authorship Verification shared tasks suggest that characterizing individual styles in long texts can be solved with almost perfect accuracy \cite{kestemont2020overview}; as a result, stylometric studies have increasingly focusing on short texts in social media or online communications for authorship profiling or verification \cite{brocardo2013authorship,vosoughi2015digital,boenninghoff2019similarity}. 
Our study points to a promising application of sociolinguistics for authorship detection in social media, as idiolect variation is evident and consistent in texts as short as 100 tokens.

\section{Learning representations of idiolects}

In forensic linguistics, textual similarity was traditionally quantified as the proportion of shared vocabulary and the number and length of shared phrases or characters (n-grams) \cite{coulthard2004author}. More sophisticated statistical methods to perform textual comparison have been developed over the years \cite{neal2017surveying,kestemont2020overview}. 

To learn representations of idiolectal style, we propose using a proxy task of authorship verification, where, given two input texts, a model must determine if they were written by the same author or not \cite{neal2017surveying}. The identification is performed by scoring the two texts under comparison with a linguistic similarity measure and, if their linguistic similarity measure exceeds a certain threshold, the two texts are judged to be written by the same author. 

\myparagraph{Task definition}
Given a collection of text pairs by multiple authors $\mathcal{X}=\{({\bf x}_1^{p_1}, {\bf x}_2^{p_1}), \dots, ({\bf x}_{n-1}^{p_{t-1}}, {\bf x}_n^{p_t})\}$ from domain $\mathcal{P}=\{p_1,p_2, \dots, p_t\}$ and labels $\mathcal{Y}=\{y_i,y_2,\dots,y_n\}$, we aim to identify a function $f_{\theta}$ that can determine whether two text samples ${\bf x}_i$ and ${\bf x}_j$ are written by the same author ($y=1$) or different authors ($y=0$). 

\myparagraph{Stylometric similarity learning}
Our model for extracting stylistic embeddings from input texts is the same as the Sentence RoBERTa or BERT network (SBERT/SRoBERTa) \cite{reimers-gurevych-2019-sentence}.  For a text pair $({\bf x}_i,{\bf x}_j)$, the Siamese model $f_{\theta}$ maps both text samples into embedding vectors $({\bf z_i}, {\bf z_j})$ in the latent space such that ${\bf z_i} = f_{\theta}({\bf x}_i)$ and ${\bf z_j} = f_{\theta}({\bf x}_j)$.  Rather than using the \texttt{[cls]} token as the representation, we use attention pooling to merge the last hidden states $[\boldsymbol{h}_0,\hdots,\boldsymbol{h}_k]$ into a single embedding vector to represent textual style.
\begin{equation}
    \begin{aligned}
        \boldsymbol{h}_{o} =& \text{AttentionPool}([\boldsymbol{h}_0,\hdots,\boldsymbol{h}_k]) \\
        \boldsymbol{z} =& \boldsymbol{W}_1\cdot\sigma(\boldsymbol{W}_2\cdot\boldsymbol{h}_{o}+\boldsymbol{b}_2)+\boldsymbol{b}_1
    \end{aligned}
\end{equation}
where $W_1$ and $W_2$ are learnable parameters and $\sigma(\cdot)$ is the ReLU activation function. Stylometric similarity between the text pair is then measured by a distance function $d({\bf z}_i,{\bf z}_j)$. Here we mainly consider the cosine similarity.
 
The underlying models are RoBERTa \cite{liu2019roberta} and BERT \cite{devlin-etal-2019-bert}. Specifically, we used the \texttt{roberta-base} or \texttt{bert-base-uncased} as the encoder.\footnote{The performance of \texttt{bert-base-cased} was almost identical to \texttt{bert-base-uncased}, so it was not included in the main results.}

\myparagraph{Loss function}
The classic max-margin loss for deep metric learning was shown to be effective in previous work on stylometry \cite{boenninghoff2019explainable,boenninghoff2019similarity}. Inspired by \citet{kim2020proxy}, we used a continuous approximation of the max-margin loss to learn the stylometric distance between users. The additional hyperparameter in this loss allows the fine-grained control of the penalty magnitude for hard samples.  

The loss function is an adaptation from the proxy-anchor loss proposed by \citet{kim2020proxy}. Given a text pair $\{{\bf x}_i, {\bf x}_j\}$, stylometric similarity between the text pair is then measured by a distance function $d({\bf z}_i,{\bf z}_j)$. Here we mainly consider the cosine similarity. To minimize the distance for same-author pairs and maximize the distance between different-author pairs, the model was trained with a contrastive loss with pre-defined margins $\{\tau_s,\tau_d\}$ for the set of positive samples $P^+$ and negative samples $P^-$ is given below. 
\begin{equation}
    \begin{aligned}
    \mathcal{L}_{(s)} = \frac{1}{|P^+|}\sum_{i,j\in P^+} \text{Softplus}\Big(\text{LogSumExp}\\\big(\alpha\cdot[d({\bf z}_i,{\bf z}_j)-\tau_s]\big)\Big)
    \end{aligned}
\end{equation}
\begin{equation}
    \begin{aligned}
    \mathcal{L}_{(s)} = \frac{1}{|P^-|}\sum_{i,j\in P^-} \text{Softplus}\Big(\text{LogSumExp}\\\big(\alpha\cdot[\tau_d-d({\bf z}_i,{\bf z}_j)]\big)\Big)
    \end{aligned}
\end{equation}
\begin{equation}
    \mathcal{L} = \mathcal{L}_{(s)} + \mathcal{L}_{(d)}
\end{equation}
where $\text{Softplus(z)}=\log(1+e^z)$ is a continuous approximation of the max function. $\alpha$ is a scaling factor that scales the penalty of the out-of-the-margin samples. The out-of-margin samples are exponentially weighted through the \textit{Log-Sum-Exp} operation such that hard examples are assigned exponentially growing weights, prompting the model to learn hard samples harder.

During inference, we compare the textual distance $d({\bf x}_1,{\bf x}_2)$ with the threshold $\tau_t$, the average of the two margins, $\tau_t=\frac{\tau_s+\tau_d}{2}$. We set $\{\tau_s=0.6,\tau_d=0.4\}$ and $\alpha=30$. 
Details about hyperparameters searching can be found in Appendix~\ref{app:hyper}.

\myparagraph{Baseline methods}
We also compare our models against several baseline methods in authorship verification: 1) {\bf GLAD}. Groningen Lightweight Authorship Detection (GLAD) \cite{hurlimann2015glad} is a binary linear classifier using multiple handcrafted linguistic features. 2) {\bf FVD}. the Feature Vector Difference (FVD) method \cite{weerasinghe:2020} is a deep learning method for authorship verification using the absolute difference between two traditional stylometric feature vectors. 3) {\bf AdHominem}. AdHominem is an LSTM-based Siamese network for authorship verification in social media domains \cite{boenninghoff2019explainable}. 4) {\bf BERT\textsubscript{Concat}/RoBERTa\textsubscript{Concat}}. The model is fine-tuned BERT/RoBERTa by concatenating two texts under comparison. Authorship was determined by performing the binary classification task on the \texttt{<cls>} token \cite{ordonez2020will}. 
Implementation details are provided in the Appendix~\ref{app:baseline}. 

\section{Data}

\myparagraph{Amazon reviews}
Our dataset was extracted from the release of the full Amazon review dataset up to 2018 \cite{ni-etal-2019-justifying}. We filtered out reviews that were shorter than 50 words to ensure sufficient text to reveal stylistic variation. We only retained users that have reviews in at least two product domains (e.g., Electronics and Books) and at least five reviews in each domain. After text cleaning, the dataset contained 128,945 users. We partitioned 40\%, 10\%, and 50\% of users into training, development, and test sets. There were 51398, 12849, and 64248 unique users in each set respectively.  As one of our goals is to analyzed stylistic variation, we reserved the majority of the data (50\% of users) for model evaluation and for subsequent linguistic analysis. The maximum length of all samples was limited to 100 tokens.

\myparagraph{Negative sampling}
For each user, we randomly sampled six pairs of texts written by the same user as positive samples of same-authorship (SA). For negative samples, we randomly sampled six texts from the rest of the data and paired them with the original text. In order to improve generalizability across domains, we enforced a sampling scheme that half of the positive/negative samples were matched in domain while the other half were cross-domain.

\myparagraph{Reddit posts} To test the generalizability, we additionally constructed a second dataset from the online community Reddit and ran a subset of experiments on it. The top 200 subreddits were extracted via the \texttt{Convokit} package \cite{chang-etal-2020-convokit}. Only users who had posted texts longer than 50 words in more than 10 subreddits were selected in the dataset, resulting in 55,368 unique users. We partitioned the 60\%, 10\%, and 30\% of users as training, development, and test sets respectively. Each user's idiolect is represented by 10 posts from 10 different subreddits. The binary labels were then generated by randomly sampling from SA and DA pairs, using the same negative sampling procedure used in the creation of the Amazon dataset.

\section{Results on proxy task}
To test whether our model does recover stylistic features, we first test its performance on the proxy task of author verification, contextualizing the performance with other models specifically designed for that task. The performance of authorship verification is evaluated by accuracy, F1 and the area-under-the-curve score (AUC) \cite{kestemont2020overview}. Table~\ref{tab:perfm_stats} suggests that all models are able to  at least  recover some distinctive aspects of individual styles even in these short text. Deep learning-based methods generally achieve better verification accuracy than GLAD \cite{hurlimann2015glad} and FVDNN \cite{weerasinghe:2020}, models based on traditional linguistic features. Siamese architecture demonstrates its usefulness in the authorship verification task, as AdHominem \cite{boenninghoff2019explainable} and SRoBERTa perform better than the pre-trained transformer RoBERTa\textsubscript{Concat}.  These results confirm that models recognize authors’ stylistic variation. Our error analysis also shows that identifying the same author across domains or different authors in the same domain poses a greater challenge to these models in general, though different model choices may exhibit various inductive biases (see Appendix~\ref{app:error_analysis}). As SRoBERTa is shown to be the most effective architecture in this task, we will use these models to examine idiolectal variation for the rest of the study.

\begin{table}[tbh]
\centering
\resizebox{0.48 \textwidth}{!}{
\begin{tabular}{lccc}
\hline
{\bf Model}  &  {\bf Accuracy} & {\bf F1} & {\bf AUC}   \\\hline
{\color{Orange} \bf Amazon reviews}\\\hline
Random & 50\% & 0.5 & 0.5 \\
GLAD &  67.1\%    &   0.667    &   0.738 \\
FVDNN & 65.1\% & 0.671 & 0.714\\
AdHominem  &  73.3\% & 0.781  &  0.811 \\\hline
BERT$_{Concat}$  & 71.3\% &  0.664 & 0.802 \\
SBERT & 76.7\% & 0.768 & 0.850\\
\hline
RoBERTa$_{Concat}$ & 73.9\% & 0.686 & 0.838   \\
SRoBERTa   &  {\bf 82.9\%} & {\bf 0.831} & {\bf 0.909} \\ \hline
{\color{PineGreen} \bf Reddit posts} \\\hline
BERT$_{Concat}$  & 65.0\% & 0.600 & 0.721 \\
SBERT & 66.3\% & 0.669 & 0.727\\
\hline
RoBERTa$_{Concat}$ & 71.0\% & 0.695 & 0.794   \\
SRoBERTa   &  {\bf 73.0\%} & {\bf 0.737} & {\bf 0.812} \\ \hline
\end{tabular}
}
\caption{Results of authorship verification on the Amazon (top) and Reddit (bottom) test samples}
\label{tab:perfm_stats}
\end{table}

\begin{table*}[tbh]
\centering
\small
\begin{tabular}{llcclcc}
\hline
\multicolumn{2}{l}{}  &                  \multicolumn{2}{c}{{\bf Amazon}} &  & \multicolumn{2}{c}{{\bf Reddit}} \\ \cline{3-4} \cline{6-7} 
\multicolumn{2}{l}{}                  & Test (Permuted)     & Test (Original)    &  & Test (Permuted)     & Test (original)    \\ \hline
\multirow{2}{*}{Lexical}   & SBERT    &  0.716 / 0.798           &   0.736 / 0.794        &  &  0.639 / 0.668            &  0.657 / 0.667         \\
                           & SRoBERTa &  0.781 / 0.856            &   0.751 / 0.850          &  &   0.654 / 0.734           &  0.686 / 0.714           \\ \hline
\multirow{2}{*}{Lexico-syntactic} & SBERT    &   0.767 / 0.851           &  0.770 / 0.852           &  &   0.681 / 0.722
           &  0.685 / 0.724           \\
                           & SRoBERTa &  {\bf 0.820 / 0.895}            &  {\bf 0.822 / 0.896}           &  & {\bf 0.730 / 0.798}             &   {\bf 0.732 / 0.800}          \\ \hline
\end{tabular}
\caption{Results (F1/AUC) on permuted examples show that models are largely insensitive to syntactic and ordering variation, and, instead, idiolect is mostly captured through lexical variation.}
\label{tab:permuted}
\end{table*}

\begin{table}[tbh]
\small
\centering
\begin{tabular}{llllll}
\hline
\multicolumn{2}{l}{}                       & \multicolumn{2}{c}{Amazon} & \multicolumn{2}{c}{Reddit} \\\hline
\multirow{2}{*}{Function} & SBERT    & \multicolumn{2}{l}{ 0.775 / 0.856}       & \multicolumn{2}{l}{0.680 / 0.744}       \\
                                & SRoBERTa & \multicolumn{2}{l}{0.786 / 0.858}       & \multicolumn{2}{l}{0.683 / 0.755}       \\ \hline
\multirow{2}{*}{Content}  & SBERT    & \multicolumn{2}{l}{0.735 / 0.812}       & \multicolumn{2}{l}{0.641 / 0.689}       \\
                                & SRoBERTa & \multicolumn{2}{l}{{\bf 0.795 / 0.870}}       & \multicolumn{2}{l}{{\bf 0.708 / 0.768}}       \\ \hline
\end{tabular}
\caption{Results (F1/AUC) show that function or content words alone are reliable authorial cues. For SRoBERTa, content words seem to convey slightly more idiolectal cues despite topical variation.}
\label{tab:function}
\end{table}

\section{Linguistic Analysis}
In this section, we seek to quantify the idiolectal variation at different linguistic levels.

\subsection{Stylometric features}
Idiolects vary at lexical, syntactic, and discourse levels, yet it remains unclear which type of variation contributes most to idiolectal variation. 

\myparagraph{Ordering} Hierarchical models of linguistic identities hold that authorial identities are reflected at all linguistic levels \cite{herring_computer-mediated_2004,grant2012txt,grant2020language}, yet the relative importance of these elements is seldom empirically explored. In order to understand the contributions of lexical distribution, syntactic ordering, or discourse coherence, we test the contributions of different linguistic features to authorship verification by perturbing the input texts. To force the model to only use lexical information, we randomly permute all tokens in our data, removing information about syntactic ordering (\texttt{lexical model}). The organization of discourse might also provide cues to idiolectal style. To test this, we preserve the word order within a sentence but permute sentences within the text to disrupt discourse information (\texttt{lecico-syntactic model}). Then we ran our experiments on these datasets using the same set of hyperparameters to compare the model performance on these perturbing inputs.

\myparagraph{Content and function words.} The use of function words has long been recognized as an important stylometric feature. A small set of function words are disproportionately frequent, relatively stable across content, and seem less under authors' conscious control \cite{kestemont-2014-function}. Yet few studies have empirically compared the relative contributions between function words and content words. To test this, we masked out all content words in the original texts with a masked token \texttt{<mask>}, which was recognized by the transformer models. For comparison, we also created masked texts with only content words. Punctuation and relative positions between words were retained as this allows the model to maximally exploit the spatial layout of content/function words.

\myparagraph{Results} 
While the importance of lexical information in authorship analysis has been emphasized, it is suggested that only using lexical information is insufficient in forensic linguistics \cite{grant2020language}. Our results in Table~\ref{tab:permuted} suggest that, even with only lexical information, the model performance is only about 4\% lower than models with access to all information. Syntactic and discourse ordering do contribute to author identities, yet the contributions are relatively minor. In forensic linguistics, it is commonly the case that only fragmentary texts are available \cite{grant2012txt}, and our findings suggest that even without broader discourse information, it is still possible to estimate author identity with good confidence. The weak contribution of discourse coherence to authorship analysis highlights that the high level organization of texts is only somewhat consistent within authors, which has been mentioned but rarely tested in forensic linguistics \cite{grant2020language}.

From Table~\ref{tab:function}, it is apparent that, even with half of the words masked out, the transformed texts still contain an abundance of reliable stylometric cues to individual writers, such that the overall accuracy is not significantly lower than models with full texts. While the importance of function words in authorship analysis has been emphasized \cite{kestemont-2014-function}, content words seem to convey slightly more idiolectal cues despite the topical variation. Both SBERT and SRoBERTa achieve similar performance on Amazon and Reddit data, yet SRoBERTa better exploits the individual variation in content words. These results strongly suggest that there are unique individual styles that are stable across topics, and our additional probing also reveals that topic information is significantly reduced in the learned embeddings (see Appendix~\ref{app:sociolect}).

\subsection{Analysis of tokenization methods}
We hypothesized that the large performance gap between BERT and RoBERTa  (\textasciitilde 5\%) could be caused by the discrepancy in the tokenization methods.
The BERT tokenizer is learned
after preprocessing the texts with heuristic rules \cite{devlin-etal-2019-bert}, whereas the BPE tokenizer for RoBERTa is learned without any additional preprocessing
or tokenization of the input \cite{liu2019roberta}. 

\myparagraph{Method} To verify, we trained several lightweight Siamese LSTM models from scratch that only differed in tokenization methods: 1) word-based tokenizer  with the vocabulary size set to either 30k or 50k to match the sizes of BPE encodings; 2) pre-trained wordpiece tokenizer for \texttt{bert-base-uncased} and \texttt{bert-base-cased}; 
3) pre-trained tokenizer for \texttt{roberta-base}. 
Implementation details are attached in Appendix~\ref{app:token}.

\begin{table}[tbh]
\centering
\small
\begin{tabular}{lccc}
\hline
{\bf Tokenization}  &  {\bf Accuracy} & {\bf F1} & {\bf AUC}   \\\hline
\texttt{Word-30k} & 67.8\% & 0.675 & 0.745 \\
\texttt{Word-50k} & 67.8\% & 0.679 & 0.744 \\
\texttt{BERT-uncased} &  67.1\%    &   0.680    &   0.737 \\
\texttt{BERT-cased} & 67.2\% & 0.677 & 0.737\\
\texttt{RoBERTa}  &  {\bf 73.1\%} & {\bf 0.734}  &  {\bf 0.804} \\\hline
\end{tabular}
\caption{The effect of tokenization methods on the model performance with respect to Amazon reviews. The pre-trained RoBERTa BPE tokenizer encodes more textual variations than the rest.}
\label{tab:tokenize}
\end{table}

\myparagraph{Results} As shown in Table~\ref{tab:tokenize}, the RoBERTa tokenizer outperforms other tokenizers by a significant margin, even though it has similar numbers of parameters to  \texttt{Word-50k}. Interestingly, pre-trained BERT tokenizer is not superior to the word-based tokenizer, despite better handling of out-of-vocabulary (OOV) tokens. 
For word-based tokenizers, increasing the vocabulary from 30k to 50k does not bring any improvements, indicating that many tokens were unused during evaluation. 
The results strongly suggest that choosing the expressive tokenizer, such as the RoBERTa BPE tokenizers directly trained on raw texts, can effectively encode more stylistic features. 
For example, for the word \texttt{cat}, the RoBERTa tokenizer gives different encodings for its common variants such as \texttt{Cat}, \texttt{CAT}, \texttt{[space]CAT} or \texttt{caaats}, but these are all  treated as the OOV tokens in word-based tokenizations. While the BERT tokenizer handles most variants, it fails to encode formatting variation such as \texttt{CAT},  \texttt{[space]CAT} and \texttt{[space][space]CAT}. Such nuances in formatting are an essential dimension of idiolectal variation in forensic analysis of electronic communications \cite{grant2012txt,grant2020language}.

\section{Characterizing idiolectal styles}
In this section, we turn our attention to distinctiveness and consistency in writing styles, both of which are key theoretical assumptions in forensic linguistics \cite{grant2012txt}.

\begin{table*}[tbh]
\begin{tabular}{p{7.5cm}p{7.5cm}}\hline
{\bf Most distinctive } & {\bf Least distinctive}  \\\hline
{\small its ok seems like a reprint i mean its not horrible but i was expecting a lil better qaulity but if i wore to do it again yes i would still buy this poster its not blurry or anything but if you have a good eye it seems a lil like a reprint }   & {\small Nice, thinner style plates that are well suited for building Lego projects. They hold Lego pieces securely and match up perfectly. Also, as a big PLUS for this company you get amazing customer service. }\\\hline            
\end{tabular}
\caption{Sample Amazon review excerpts with the most and the least distinctive style as predicted by SRoBERTa.}
\label{tab:sample}
\end{table*}

\myparagraph{Distinctiveness}
We examine inter-author variation through {\bf inter-author distinctiveness} by constructing a graph that connects users with similar style embeddings, described next. For each user in the test set, we randomly sampled one text sample and extracted its embedding through the Siamese models. Then we created the pairwise similarity matrix $M$ by computing the pairwise similarity between each text pair. Then $M$ is pruned by removed entries below a threshold $\tau_{cutoff}$, the same threshold $\tau_t$ that is used to determined SA or DA pairs. The pruned matrix $\hat{M}$ is treated as the graph adjacency matrix from which a network $G$ is constructed. 
\begin{equation}
    S_i = 1 - \frac{\sum_j^N\mathbb{I}[j\in V_i]}{N}
\end{equation}
where $V_i$ is the set of neighbors of node $i$ in G, $N$ the total node count, and ${\mathbb{I}}$[ ] the indicator function. $\sum_j^N\mathbb{I}[j\in V_i]$ is the degree centrality of node $i$. We found that features from the unweighted graph are perfectly correlated with the ones from the weighted graph. The unweighted graph is kept for computational efficiency. The scores were averaged over 5 runs. The intuition is that, since authors are connected to similar authors, the more neighbors an author has, the less distinctive their style is. A distinctiveness of 0.6 implies that this author is different from 60\% of authors in the dataset.

\myparagraph{Consistency}
We also measured the {\bf intra-author consistency} in styles through concentration in the latent space. The concentration can be quantified by the conicity, a measure of the averaged vector alignment to mean (ATM), as in the following equation \cite{chandrahas-etal-2018-towards}.
\begin{equation}
    Conicity({\bf V}) = \frac{1}{|{\bf V}|}\sum_{{\bf v}\in{\bf V}}ATM({\bf v},{\bf V})
\end{equation}
\begin{equation}
    ATM({\bf v},{\bf V}) = cosine\Big({\bf v},\frac{1}{|{\bf V}|}\sum_{{\bf x}\in{\bf V}}{\bf x}\Big)
\end{equation}
where ${\bf v}\in{\bf V}$ is a latent vector in the set of vectors ${\bf V}$. The ATM measures the cosine distance between ${\bf v}$ to the centroid of ${\bf V}$ whereas the conicity indicates the overall clusteredness of vectors in ${\bf V}$ around the centroid. If all texts written by the same user are highly aligned around their centroid with a conicity close to 1, this suggests that this user is highly consistent in writing style.

\begin{figure}
    \centering
    \includegraphics[width=\linewidth]{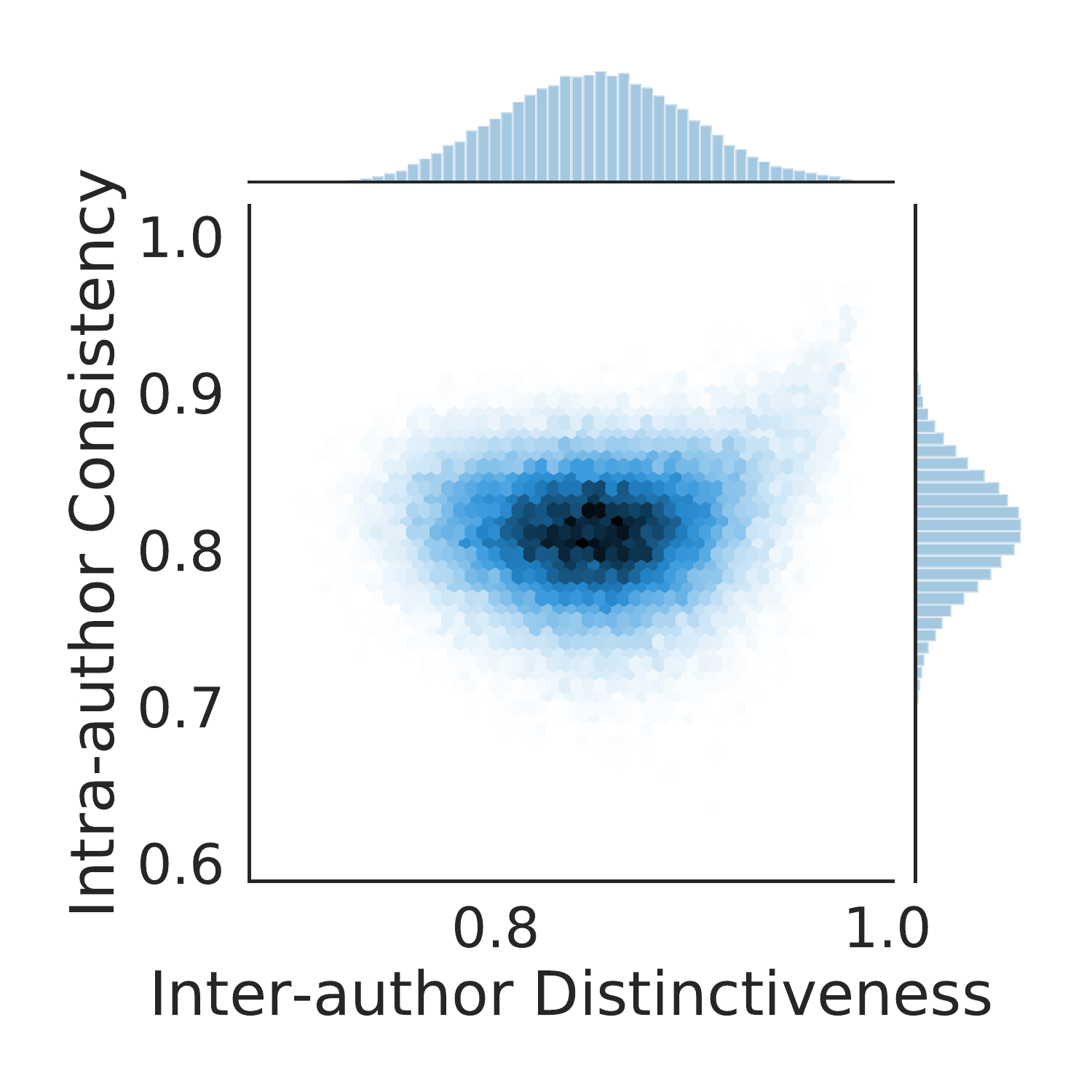}
    \caption{The joint distribution of distinctiveness and consistency on the Amazon reviews as computed with SRoBERTa embeddings. Both follow normal distributions yet there is no meaningful correlation between them, suggesting that these two dimensions may vary independently of each other.}
    \label{fig:joint}
\end{figure}

\myparagraph{Analysis}
The distributions of style distinctiveness and consistency both conform to a normal distribution (Figure~\ref{fig:joint}), yet no meaningful correlation exists between these two measures (\texttt{Amazon}: Spearman's $\rho$=0.078; \texttt{Reddit}: Spearman's $\rho$=0.11). In general, users are highly consistent in their writing styles even in such a large population, with an average of 0.8, much higher than that for random samples (\textasciitilde0.4). Users are also quite distinctive from one another, as on average a user's style is different from 80\% of users in the population pool. Yet individuals do differ in their degrees of distinctiveness and consistency, which may be taken into consideration in forensic linguistics. Because inconsistency or indistinctiveness may weaken the linguistic evidence to be analyzed.

In Table~\ref{tab:sample}, the least distinctive text is characterized by plain language, proper formatting, and typical content, which reflects the unmarked style of stereotypical Amazon reviews. Yet this review itself is still quite distinctive as it differs from 60\% of the total reviews. The most distinctive review exhibits multiple deviations from the norm of this genre. The style is unconventional with uncapitalized letters, run-on sentences, typos, the lack of periods, and the use of colloquial alternative spellings such as ``haft", ``lil" and ``wore", all of which make this review highly marked. For style consistency, the most consistent writers incline towards using similar formatting, emojis, and narrative perspectives across reviews, whereas the least consistent users tend to shift across registers and perspectives in writings (see Appendix~\ref{app:dc} for additional samples).

We tested how various authors affect the verification performance. To avoid circular validation resulting from repeatedly using the same training data, we retrained the model with the repartitioned test data and tested them using the development set. The original test set was repartitioned into three disjoint chunks of equal size, each chunk containing authors solely from either the top, middle or bottom 33\% in terms of distinctiveness or consistency. Results in Table~\ref{tab:partition} suggest that, while most models performed similarly, models trained on inconsistent or indistinctive authors significantly underperformed. This result may have implications for comparative authorship analysis in that it is desirable to control the number of inconsistent or indistinctive authors in the dataset.

\begin{table}[]
\small
\begin{tabular}{clll}
\hline
\multicolumn{1}{l}{}             & Range    & Amazon & Reddit \\ \hline
\multirow{4}{*}{Consistent}     & Random   &  {\bf 0.806} / 0.875      &  0.677 / 0.761      \\
                                 & High     &  0.801 / {\bf 0.879}      & {\bf 0.680} / {\bf 0.767}       \\
                                 & Moderate & 0.805 / 0.874       & 0.672 / 0.757       \\
                                 & Low      & 0.797 / 0.867       &  0.661 / 0.716      \\ \hline
\multirow{4}{*}{Distinctive} & Random   &   0.808 / 0.876     &   0.695 / {\bf 0.754}     \\
                                 & High     &  0.803 / 0.874      &  {\bf 0.709} / 0.750      \\
                                 & Moderate &  {\bf 0.808} / {\bf 0.880}      &  0.663 / 0.731      \\
                                 & Low      &  0.793 / 0.861      & 0.611 / 0.68       \\ \hline
\end{tabular}
\caption{Performance (F1/AUC) on different partitions of data. Models trained on inconsistent or indistinctive authors significantly underperformed.}
\label{tab:partition}
\end{table}

\section{Compositionality of styles}

Finally, we sought to understand how stylistic variations are encoded. At least for certain stylistic features, there is {\it additive stylistic compositionality} in the latent space onto which the texts are projected (Figure~\ref{fig:composition}). 

\begin{table}[]
\centering
\small
\begin{tabular}{lrr}
\hline
{\bf Style Shift} & {\bf Amazon} & {\bf Reddit} \\ \hline
Random      &    0.032      &   0.002    \\ \hline
and $\rightarrow$ \&       &    0.698     &   0.484   \\
. $\rightarrow$ space    &   0.389       &   0.352    \\
. $\rightarrow$ !!!!           &    0.569      &  0.396\\
lower-cased & 0.679 & 0.660 \\
upper-cased & 0.463 & 0.470 \\
I $\rightarrow$ $\varnothing$ & 0.420 & 0.306      \\ 
going to $\rightarrow$ gonna & 0.398 & 0.390 \\
want to $\rightarrow$ wanna & 0.478 & 0.505\\
-ing $\rightarrow$ -in' & 0.522 & 0.484 \\
w/o elongation $\rightarrow$ w/ & 0.410 & 0.343  \\
\hline
\end{tabular}
\caption{ $\overline{S_{ij}}$ for qualitative style shifts, averaged across all comparisons. The offset vectors for style shifts are highly aligned.}
\label{tab:analogy}
\end{table}

\begin{table}[]
\small
\centering
\begin{tabular}{lrrrr}
\hline
\multirow{2}{*}{{\bf Style shift}} & \multicolumn{2}{c}{{\bf Amazon}} & \multicolumn{2}{c}{{\bf Reddit}} \\\cline{2-5}
     & \addstackgap{$\overline{S_k}$}          & $\overline{\Delta Norm}$          & $\overline{S_k}$         &$\overline{\Delta Norm}$  \\\hline
Random & 0.686 & 0.059 & 0.692 & 0.059 \\
and $\rightarrow$ \&  & 0.957 & 0.176 & 0.922 & 0.124\\
. $\rightarrow$ !!!!     & 0.964   & 0.085 & 0.953 & 0.072\\
I $\rightarrow$ $\varnothing$    &  0.860 & 0.257   & 0.831 & 0.273\\
-ing $\rightarrow$ -in' & 0.933 & 0.103 & 0.928 & 0.097\\\hline
\end{tabular}
\caption{$\overline{S_k}$ and $\overline{\Delta Norm}$ for quantitative style shifts. The direction and the magnitude of change encode the type and the degree of style shift. $\overline{\Delta Norm}$ is positive, suggesting that more manipulations result in a longer offset vector along that direction.}
\label{tab:quant}
\end{table}

\myparagraph{Method} 
In light of the word analogy task for word embeddings \cite{mikolov-etal-2013-linguistic,linzen-2016-issues}, we designed a series of linguistic stimuli that vary systematically in styles to probe the structure of the stylistic embeddings. For each stylistic dimension, we created $n$ text embedding pairs $\mathcal{P}=[({\bf p}_r^1,{\bf p}_m^1), \dots, ({\bf p}_r^n,{\bf p}_m^n)]$ where ${\bf p}_r^i$ is the embedding of a randomly sampled text and ${\bf p}_m^i$ is the embedding of the modified version of ${\bf p}_r^i$ so that it differs from ${\bf p}_r^i$ in only one stylistic aspect. For sample $i$ and $j$ from $\mathcal{P}$, we quantified
\begin{equation}
    S_{ij} = cosine({\bf p}_r^i-{\bf p}_m^i,{\bf p}_r^j-{\bf p}_m^j)
\end{equation}
Like the word analogy task, if this stylistic dimension is encoded only in one direction, we should expect $S_{ij}$ close to 1.

For a target qualitative style shift, we randomly sampled 1000 texts and modified the text to approximate the target stylistic dimension. For example, if null subject is the target feature, we remove the subjective ``I'' from ``I recommend crispy pork!" to ``Recommend crispy pork!". Then we compute $S_{ij}$ for each pair, totaling 499500 possible comparisons. Here we selected 10 stylistic markers of textual formality for evaluation \cite{macleodwhose,biber2019register} and these textual modifications cover insertion, deletion, and replacement operations.  

For quantitative style shifts, we measure $S_{k}$ between samples as well as the difference in length with the following equation. We compare two embedding pairs $({\bf p}_r^k,{\bf p}_s^k)$ and $({\bf p}_r^k,{\bf p}_l^k)$, where both ${\bf p}_s^k$ and ${\bf p}_l^k$ differ from ${\bf p}_r^i$ in only one stylistic dimension.  But ${\bf p}_l^k$ is further along that dimension than ${\bf p}_s^k$. For instance, compared to the original review ${\bf p}_r^k$, ${\bf p}_s^k$ contains five more ``!!!" whereas ${\bf p}_l^k$ contains ten more such tokens. Here we surveyed four stylistic markers of formality. 
\begin{equation}
    \begin{aligned}
    S_{k} =&  cosine({\bf p}_l^k-{\bf p}_r^k,{\bf p}_s^k-{\bf p}_r^k) \\
    \Delta norm_k &= ||{\bf p}_l^k-{\bf p}_r^k||_2 - ||{\bf p}_s^k-{\bf p}_r^k||_2
    \end{aligned}
\end{equation}
For each stylistic shift, we collected 2000 samples, each contained at least 8 markers of that style. Then we modified the original review ${\bf p}_r^k$ to the target style by incrementally transforming the keywords into the target keywords by 50\% (${\bf p}_s^k$) and 100\% (${\bf p}_l^k$). If these styles are highly organized, we should expect $S_{k}$ to be close to 1, suggesting that changes in the same dimension point to the same direction. Yet we also expect that quantitative changes should also be reflected in the significant length difference (magnitude of $\Delta norm_k$) and the direction of the difference ($\Delta norm_k$ being positive or negative) of the style vectors. 

\myparagraph{Results}
Results in Table~\ref{tab:analogy} suggest that both models outperform the random baseline, $S_{ij}$ generated with the same samples by randomly replacing some words. Like word embeddings, stylistic embeddings also exhibit a linear additive relationship between various stylistic attributes.
In Figure~\ref{fig:composition}, converting all letters to lower case causes textual representations to move collectively in approximately the same direction. Despite such regularities, the offset vectors for style shifts were not perfectly aligned in all instances, which may be attributed to the variations across texts. 

For quantitative changes, SRoBERTa on both Amazon and Reddit data encode the same type of change in the same direction, as the vector offsets are highly aligned to each other (Table~\ref{tab:quant}). Yet, greater degrees of style shift relative to the original text translate to a larger magnitude of change along that direction ( $\Delta norm_k$ in Table~\ref{tab:quant}). In Figure~\ref{fig:composition}, after removing the first three instances or all the occurrences of ``I" from the original text, the resulted representations both shift in the same direction but differ in magnitude. Such changes cannot be explained by random variation, suggesting that both models learn to encode fine-grained stylistic dimensions in the latent space through the proxy task. 

While we only examine several stylistic markers, we were aware that the learned style representations also exhibit regularities for other lexical manipulations, as long as the manipulation is systematic and regular across samples. An explanation is that the model is systematically tracking the fine-grained variations in lexical statistics. Yet the proposed model must also encode more abstract linguistic features because it outperformed GLAD \cite{hurlimann2015glad} and FVDNN \cite{weerasinghe:2020} that also track bag-of-words or bag-of-ngram features. Previous research in word embeddings attributes the performance of the analogy task to the occurrence patterns \cite{pennington-etal-2014-glove,levy-etal-2015-improving,ethayarajh-etal-2019-understanding}. The fact that these variations are encoded systematically beyond random variation and at such a fine-grain manner indicates that they are stylistic dimensions along which individual choices vary frequently and regularly.

\section{Conclusions}

The relatively unconstrained nature of the online genres tolerates a much wider range of stylistic variation than conventional genres \cite{hovy2015user}. Online genres are often marked by unconventional spellings, heavy use of colloquial language, extensive deviations in formatting, and the relaxation of grammatical rules, providing rich linguistic resources to construct and perform one's identity.
Our analysis of idiolects in online registers has highlighted that idiolectal variations permeate all linguistic levels, present in both surface lexico-syntactic features and high-level discourse organization. Traditional sociolinguistic research often regards idiolects as idiosyncratic and unstable and not as regular as sociolects \cite{labov1989exact,barlow2018individual}; here, we show that idiolectal variation is not only highly distinctive but also consistent, even in a relatively large population. Our findings suggest that individuals may differ considerably by degrees of consistency and distinctiveness across multiple text samples, which sheds light on the theoretical discussions and practical applications in forensic linguistics. Our findings also have implications for sociolinguistics, as we have shown an effective method to discover, understand and exploit sociolinguistic variation.

\section{Ethical considerations}
While this study is theoretically driven, we are aware that there might be some ethical concerns for models surveyed in this study. While computational stylometry can be applied to forensic investigations \cite{grant2020language}, authorship verification in online social networks, if put to malicious use, may weaken the anonymity of some users, leading to potential privacy issues. Our results also show that caution should be taken when deploying these models in forensic scenarios, as different models or tokenizers might show different inductive biases that may bias towards certain types of users. Another potential bias is that we only selected a small group of the most productive writers from the pool (less than 20\% of all data), but this sample might not necessary represent all populations. We urge that caution should be exercised when using these models in real-life settings. 

We still consider that the benefits of our study outweigh potential dangers. Deep learning-based stylometry is an active research area in recent years. While many studies focus on improving performance, we provide insights into how some of these models make decisions based on certain linguistic features and expose some of the models' inductive biases. This interpretability analysis could be used to guide the proper use of these methods in decision-making processes. The analysis could also be useful in developing adversarial techniques that guard against the malicious use of such technologies. 

All of our experiments were performed on public data, in accordance with the terms of service. All authors were anonymized. In addition, the term ``Siamese”  may be considered offensive when it refers to some groups of people. The use of this term follows the research naming of a mathematical model in machine learning literature. We use the word here to refer to a neural network model and make no reference to particular population groups. 

\section*{Acknowledgements}
We thank Professor Patrice Beddor, Jiaxin Pei at the University of Michigan and Zuoyu Tian at the University of Indiana Bloomington for their helpful discussions. We are also grateful for comments from anonymous reviewers, which helped improve the paper greatly. This material is based upon work supported by the National Science Foundation under Grant No 1850221.

\bibliography{anthology,custom}

\begin{thebibliography}{52}
\expandafter\ifx\csname natexlab\endcsname\relax\def\natexlab#1{#1}\fi

\bibitem[{Barlow(2013)}]{barlow2013individual}
Michael Barlow. 2013.
\newblock Individual differences and usage-based grammar.
\newblock \emph{International Journal of Corpus Linguistics}, 18(4):443--478.

\bibitem[{Barlow(2018)}]{barlow2018individual}
Michael Barlow. 2018.
\newblock The individual and the group from a corpus perspective.
\newblock \emph{The Corpus Linguistic Discourse: In Honour of Wolfgang Teubert.
  Amsterdam}, pages 163--184.

\bibitem[{Basile et~al.(2019)Basile, Gatt, and Nissim}]{basile-etal-2019-write}
Angelo Basile, Albert Gatt, and Malvina Nissim. 2019.
\newblock \href {https://doi.org/10.18653/v1/P19-1246} {You write like you eat:
  Stylistic variation as a predictor of social stratification}.
\newblock In \emph{Proceedings of the 57th Annual Meeting of the Association
  for Computational Linguistics}, pages 2583--2593, Florence, Italy.
  Association for Computational Linguistics.

\bibitem[{Belainine et~al.(2020)Belainine, Sadat, Boukadoum, and
  Lounis}]{belainine-etal-2020-towards}
Billal Belainine, Fatiha Sadat, Mounir Boukadoum, and Hakim Lounis. 2020.
\newblock \href {https://www.aclweb.org/anthology/2020.lincr-1.7} {Towards a
  multi-dataset for complex emotions learning based on deep neural networks}.
\newblock In \emph{Proceedings of the Second Workshop on Linguistic and
  Neurocognitive Resources}, pages 50--58, Marseille, France. European Language
  Resources Association.

\bibitem[{Biber and Conrad(2019)}]{biber2019register}
Douglas Biber and Susan Conrad. 2019.
\newblock \emph{Register, Genre, and Style}.
\newblock Cambridge University Press.

\bibitem[{Bloch(1948)}]{bloch1948set}
Bernard Bloch. 1948.
\newblock A set of postulates for phonemic analysis.
\newblock \emph{Language}, 24(1):3--46.

\bibitem[{Boenninghoff et~al.(2019{\natexlab{a}})Boenninghoff, Hessler,
  Kolossa, and Nickel}]{boenninghoff2019explainable}
Benedikt Boenninghoff, Steffen Hessler, Dorothea Kolossa, and Robert~M Nickel.
  2019{\natexlab{a}}.
\newblock Explainable authorship verification in social media via
  attention-based similarity learning.
\newblock In \emph{2019 IEEE International Conference on Big Data}, pages
  36--45. IEEE.

\bibitem[{Boenninghoff et~al.(2019{\natexlab{b}})Boenninghoff, Nickel, Zeiler,
  and Kolossa}]{boenninghoff2019similarity}
Benedikt Boenninghoff, Robert~M Nickel, Steffen Zeiler, and Dorothea Kolossa.
  2019{\natexlab{b}}.
\newblock Similarity learning for authorship verification in social media.
\newblock In \emph{ICASSP 2019-2019 IEEE International Conference on Acoustics,
  Speech and Signal Processing (ICASSP)}, pages 2457--2461. IEEE.

\bibitem[{Brocardo et~al.(2013)Brocardo, Traore, Saad, and
  Woungang}]{brocardo2013authorship}
Marcelo~Luiz Brocardo, Issa Traore, Sherif Saad, and Isaac Woungang. 2013.
\newblock Authorship verification for short messages using stylometry.
\newblock In \emph{2013 International Conference on Computer, Information and
  Telecommunication Systems (CITS)}, pages 1--6. IEEE.

\bibitem[{{Chandrahas} et~al.(2018){Chandrahas}, Sharma, and
  Talukdar}]{chandrahas-etal-2018-towards}
{Chandrahas}, Aditya Sharma, and Partha Talukdar. 2018.
\newblock \href {https://doi.org/10.18653/v1/P18-1012} {Towards understanding
  the geometry of knowledge graph embeddings}.
\newblock In \emph{Proceedings of the 56th Annual Meeting of the Association
  for Computational Linguistics (Volume 1: Long Papers)}, pages 122--131,
  Melbourne, Australia. Association for Computational Linguistics.

\bibitem[{Chang et~al.(2020)Chang, Chiam, Fu, Wang, Zhang, and
  Danescu-Niculescu-Mizil}]{chang-etal-2020-convokit}
Jonathan~P. Chang, Caleb Chiam, Liye Fu, Andrew Wang, Justine Zhang, and
  Cristian Danescu-Niculescu-Mizil. 2020.
\newblock \href {https://www.aclweb.org/anthology/2020.sigdial-1.8}
  {{C}onvo{K}it: A toolkit for the analysis of conversations}.
\newblock In \emph{Proceedings of the 21th Annual Meeting of the Special
  Interest Group on Discourse and Dialogue}, pages 57--60, 1st virtual meeting.
  Association for Computational Linguistics.

\bibitem[{Coulthard(2004)}]{coulthard2004author}
Malcolm Coulthard. 2004.
\newblock Author identification, idiolect, and linguistic uniqueness.
\newblock \emph{Applied Linguistics}, 25(4):431--447.

\bibitem[{Coulthard et~al.(2016)Coulthard, Johnson, and
  Wright}]{coulthard2016introduction}
Malcolm Coulthard, Alison Johnson, and David Wright. 2016.
\newblock \emph{An introduction to forensic linguistics: Language in evidence}.
\newblock Routledge.

\bibitem[{Devlin et~al.(2019)Devlin, Chang, Lee, and
  Toutanova}]{devlin-etal-2019-bert}
Jacob Devlin, Ming-Wei Chang, Kenton Lee, and Kristina Toutanova. 2019.
\newblock \href {https://doi.org/10.18653/v1/N19-1423} {{BERT}: Pre-training of
  deep bidirectional transformers for language understanding}.
\newblock In \emph{Proceedings of the 2019 Conference of the North {A}merican
  Chapter of the Association for Computational Linguistics: Human Language
  Technologies, Volume 1 (Long and Short Papers)}, pages 4171--4186,
  Minneapolis, Minnesota. Association for Computational Linguistics.

\bibitem[{Eckert(2012)}]{eckert2012three}
Penelope Eckert. 2012.
\newblock Three waves of variation study: The emergence of meaning in the study
  of sociolinguistic variation.
\newblock \emph{Annual review of Anthropology}, 41:87--100.

\bibitem[{Ethayarajh et~al.(2019)Ethayarajh, Duvenaud, and
  Hirst}]{ethayarajh-etal-2019-understanding}
Kawin Ethayarajh, David Duvenaud, and Graeme Hirst. 2019.
\newblock \href {https://doi.org/10.18653/v1/P19-1166} {Understanding
  undesirable word embedding associations}.
\newblock In \emph{Proceedings of the 57th Annual Meeting of the Association
  for Computational Linguistics}, pages 1696--1705, Florence, Italy.
  Association for Computational Linguistics.

\bibitem[{Flekova et~al.(2016)Flekova, Preo{\c{t}}iuc-Pietro, and
  Ungar}]{flekova-etal-2016-exploring}
Lucie Flekova, Daniel Preo{\c{t}}iuc-Pietro, and Lyle Ungar. 2016.
\newblock \href {https://doi.org/10.18653/v1/P16-2051} {Exploring stylistic
  variation with age and income on {T}witter}.
\newblock In \emph{Proceedings of the 54th Annual Meeting of the Association
  for Computational Linguistics (Volume 2: Short Papers)}, pages 313--319,
  Berlin, Germany. Association for Computational Linguistics.

\bibitem[{Grant(2012)}]{grant2012txt}
Tim Grant. 2012.
\newblock Txt 4n6: method, consistency, and distinctiveness in the analysis of
  sms text messages.
\newblock \emph{Journal of Law and Policy}, 21:467.

\bibitem[{Grant and MacLeod(2020)}]{grant2020language}
Tim Grant and Nicci MacLeod. 2020.
\newblock \emph{Language and Online Identities: The Undercover Policing of
  Internet Sexual Crime}.
\newblock Cambridge University Press.

\bibitem[{Grant and MacLeod(2018)}]{grant2018resources}
Tim Grant and Nicola MacLeod. 2018.
\newblock Resources and constraints in linguistic identity performance--a
  theory of authorship.
\newblock \emph{Language and Law/Linguagem e Direito}, 5(1):80--96.

\bibitem[{Herring(2004)}]{herring_computer-mediated_2004}
Susan~C. Herring. 2004.
\newblock Computer-{Mediated} {Discourse} {Analysis}: {An} {Approach} to
  {Researching} {Online} {Behavior}.
\newblock Learning in doing, pages 338--376. Cambridge University Press, New
  York, NY, US.

\bibitem[{Holmes(1998)}]{holmes1998evolution}
David~I Holmes. 1998.
\newblock The evolution of stylometry in humanities scholarship.
\newblock \emph{Literary and linguistic computing}, 13(3):111--117.

\bibitem[{Hovy et~al.(2015)Hovy, Johannsen, and S{\o}gaard}]{hovy2015user}
Dirk Hovy, Anders Johannsen, and Anders S{\o}gaard. 2015.
\newblock User review sites as a resource for large-scale sociolinguistic
  studies.
\newblock In \emph{Proceedings of the 24th international conference on World
  Wide Web}, pages 452--461.

\bibitem[{H{\"u}rlimann et~al.(2015)H{\"u}rlimann, Weck, van~den Berg, Suster,
  and Nissim}]{hurlimann2015glad}
Manuela H{\"u}rlimann, Benno Weck, Esther van~den Berg, Simon Suster, and
  Malvina Nissim. 2015.
\newblock Glad: Groningen lightweight authorship detection.
\newblock In \emph{CLEF (Working Notes)}.

\bibitem[{Johnson and Wright(2014)}]{johnson2017identifying}
Alison Johnson and David Wright. 2014.
\newblock Identifying idiolect in forensic authorship attribution: an n-gram
  textbite approach.
\newblock \emph{Language and Law}, 1(1).

\bibitem[{Johnstone and Bean(1997)}]{johnstone1997self}
Barbara Johnstone and Judith~Mattson Bean. 1997.
\newblock Self-expression and linguistic variation.
\newblock \emph{Language in Society}, 26(2):221--246.

\bibitem[{Kestemont(2014)}]{kestemont-2014-function}
Mike Kestemont. 2014.
\newblock \href {https://doi.org/10.3115/v1/W14-0908} {Function words in
  authorship attribution. from black magic to theory?}
\newblock In \emph{Proceedings of the 3rd Workshop on Computational Linguistics
  for Literature ({CLFL})}, pages 59--66, Gothenburg, Sweden. Association for
  Computational Linguistics.

\bibitem[{Kestemont et~al.(2020)Kestemont, Manjavacas, Markov, Bevendorff,
  Wiegmann, Stamatatos, Potthast, and Stein}]{kestemont2020overview}
Mike Kestemont, Enrique Manjavacas, Ilia Markov, Janek Bevendorff, Matti
  Wiegmann, Efstathios Stamatatos, Martin Potthast, and Benno Stein. 2020.
\newblock Overview of the cross-domain authorship verification task at pan
  2020.
\newblock In \emph{CLEF}.

\bibitem[{Kestemont et~al.(2019)Kestemont, Stamatatos, Manjavacas, Daelemans,
  Potthast, and Stein}]{kestemont:2019}
Mike Kestemont, Efstathios Stamatatos, Enrique Manjavacas, Walter Daelemans,
  Martin Potthast, and Benno Stein. 2019.
\newblock \href {http://ceur-ws.org/Vol-2380/} {{Overview of the Cross-domain
  Authorship Attribution Task at PAN 2019}}.
\newblock In \emph{{CLEF 2019 Labs and Workshops, Notebook Papers}}.
  CEUR-WS.org.

\bibitem[{Kestemont et~al.(2018)Kestemont, Tschuggnall, Stamatatos, Daelemans,
  Specht, Stein, and Potthast}]{kestemont:2018}
Mike Kestemont, Michael Tschuggnall, Efstathios Stamatatos, Walter Daelemans,
  G{\"u}nther Specht, Benno Stein, and Martin Potthast. 2018.
\newblock \href {http://ceur-ws.org/Vol-2125/} {{Overview of the Author
  Identification Task at PAN-2018: Cross-domain Authorship Attribution and
  Style Change Detection}}.
\newblock In \emph{Working Notes Papers of the CLEF 2018 Evaluation Labs},
  volume 2125 of \emph{CEUR Workshop Proceedings}. CEUR-WS.org.

\bibitem[{Kim et~al.(2020)Kim, Kim, Cho, and Kwak}]{kim2020proxy}
Sungyeon Kim, Dongwon Kim, Minsu Cho, and Suha Kwak. 2020.
\newblock Proxy anchor loss for deep metric learning.
\newblock In \emph{Proceedings of the IEEE/CVF Conference on Computer Vision
  and Pattern Recognition}, pages 3238--3247.

\bibitem[{Labov(1972)}]{labov1972sociolinguistic}
William Labov. 1972.
\newblock \emph{Sociolinguistic patterns}.
\newblock 4. University of Pennsylvania Press.

\bibitem[{Labov(1989)}]{labov1989exact}
William Labov. 1989.
\newblock Exact description of the speech community: Short a in philadelphia.
\newblock \emph{Language Change and Variation}, pages 1--57.

\bibitem[{Levy et~al.(2015)Levy, Goldberg, and
  Dagan}]{levy-etal-2015-improving}
Omer Levy, Yoav Goldberg, and Ido Dagan. 2015.
\newblock \href {https://doi.org/10.1162/tacl_a_00134} {Improving
  distributional similarity with lessons learned from word embeddings}.
\newblock \emph{Transactions of the Association for Computational Linguistics},
  3:211--225.

\bibitem[{Linzen(2016)}]{linzen-2016-issues}
Tal Linzen. 2016.
\newblock \href {https://doi.org/10.18653/v1/W16-2503} {Issues in evaluating
  semantic spaces using word analogies}.
\newblock In \emph{Proceedings of the 1st Workshop on Evaluating Vector-Space
  Representations for {NLP}}, pages 13--18, Berlin, Germany. Association for
  Computational Linguistics.

\bibitem[{Liu et~al.(2019)Liu, Ott, Goyal, Du, Joshi, Chen, Levy, Lewis,
  Zettlemoyer, and Stoyanov}]{liu2019roberta}
Yinhan Liu, Myle Ott, Naman Goyal, Jingfei Du, Mandar Joshi, Danqi Chen, Omer
  Levy, Mike Lewis, Luke Zettlemoyer, and Veselin Stoyanov. 2019.
\newblock Roberta: A robustly optimized bert pretraining approach.
\newblock \emph{arXiv preprint arXiv:1907.11692}.

\bibitem[{MacLeod and Grant(2012)}]{macleodwhose}
Nicci MacLeod and Tim Grant. 2012.
\newblock Whose tweet? authorship analysis of micro-blogs and other short-form
  messages.
\newblock In \emph{Proceedings of The International Association of Forensic
  Linguists’ Tenth Biennial Conference}.

\bibitem[{Meyerhoff and Walker(2007)}]{meyerhoff2007persistence}
Miriam Meyerhoff and James~A Walker. 2007.
\newblock The persistence of variation in individual grammars: Copula absence
  in ‘urban sojourners’ and their stay-at-home peers, bequia (st vincent
  and the grenadines) 1.
\newblock \emph{Journal of Sociolinguistics}, 11(3):346--366.

\bibitem[{Mikolov et~al.(2013)Mikolov, Yih, and
  Zweig}]{mikolov-etal-2013-linguistic}
Tomas Mikolov, Wen-tau Yih, and Geoffrey Zweig. 2013.
\newblock \href {https://www.aclweb.org/anthology/N13-1090} {Linguistic
  regularities in continuous space word representations}.
\newblock In \emph{Proceedings of the 2013 Conference of the North {A}merican
  Chapter of the Association for Computational Linguistics: Human Language
  Technologies}, pages 746--751, Atlanta, Georgia. Association for
  Computational Linguistics.

\bibitem[{Neal et~al.(2017)Neal, Sundararajan, Fatima, Yan, Xiang, and
  Woodard}]{neal2017surveying}
Tempestt Neal, Kalaivani Sundararajan, Aneez Fatima, Yiming Yan, Yingfei Xiang,
  and Damon Woodard. 2017.
\newblock Surveying stylometry techniques and applications.
\newblock \emph{ACM Computing Surveys (CSUR)}, 50(6):1--36.

\bibitem[{Ni et~al.(2019)Ni, Li, and McAuley}]{ni-etal-2019-justifying}
Jianmo Ni, Jiacheng Li, and Julian McAuley. 2019.
\newblock \href {https://doi.org/10.18653/v1/D19-1018} {Justifying
  recommendations using distantly-labeled reviews and fine-grained aspects}.
\newblock In \emph{Proceedings of the 2019 Conference on Empirical Methods in
  Natural Language Processing and the 9th International Joint Conference on
  Natural Language Processing (EMNLP-IJCNLP)}, pages 188--197, Hong Kong,
  China. Association for Computational Linguistics.

\bibitem[{Ordo{\~n}ez et~al.(2020)Ordo{\~n}ez, Soto, and
  Chen}]{ordonez2020will}
Juanita Ordo{\~n}ez, Rafael~Rivera Soto, and Barry~Y Chen. 2020.
\newblock Will longformers pan out for authorship verification.
\newblock \emph{Working Notes of CLEF}.

\bibitem[{Pennington et~al.(2014)Pennington, Socher, and
  Manning}]{pennington-etal-2014-glove}
Jeffrey Pennington, Richard Socher, and Christopher Manning. 2014.
\newblock \href {https://doi.org/10.3115/v1/D14-1162} {{G}lo{V}e: Global
  vectors for word representation}.
\newblock In \emph{Proceedings of the 2014 Conference on Empirical Methods in
  Natural Language Processing ({EMNLP})}, pages 1532--1543, Doha, Qatar.
  Association for Computational Linguistics.

\bibitem[{Reimers and Gurevych(2019)}]{reimers-gurevych-2019-sentence}
Nils Reimers and Iryna Gurevych. 2019.
\newblock \href {https://doi.org/10.18653/v1/D19-1410} {Sentence-{BERT}:
  Sentence embeddings using {S}iamese {BERT}-networks}.
\newblock In \emph{Proceedings of the 2019 Conference on Empirical Methods in
  Natural Language Processing and the 9th International Joint Conference on
  Natural Language Processing (EMNLP-IJCNLP)}, pages 3982--3992, Hong Kong,
  China. Association for Computational Linguistics.

\bibitem[{Schilling-Estes(1998)}]{schilling1998investigating}
Natalie Schilling-Estes. 1998.
\newblock Investigating “self-conscious” speech: The performance register
  in ocracoke english.
\newblock \emph{Language in society}, 27(1):53--83.

\bibitem[{Turell(2010)}]{turell2010use}
M~Teresa Turell. 2010.
\newblock The use of textual, grammatical and sociolinguistic evidence in
  forensic text comparison.
\newblock \emph{International Journal of Speech, Language \& the Law}, 17(2).

\bibitem[{Vosoughi et~al.(2015)Vosoughi, Zhou, and Roy}]{vosoughi2015digital}
Soroush Vosoughi, Helen Zhou, and Deb Roy. 2015.
\newblock Digital stylometry: Linking profiles across social networks.
\newblock In \emph{International Conference on Social Informatics}, pages
  164--177. Springer.

\bibitem[{Wardhaugh(2011)}]{wardhaugh2011introduction}
Ronald Wardhaugh. 2011.
\newblock \emph{An introduction to sociolinguistics}, volume~28.
\newblock John Wiley \& Sons.

\bibitem[{Weerasinghe and Greenstadt(2020)}]{weerasinghe:2020}
Janith Weerasinghe and Rachel Greenstadt. 2020.
\newblock \href
  {https://pan.webis.de/downloads/publications/papers/weerasinghe_2020.pdf}
  {{Feature Vector Difference based Neural Network and Logistic Regression
  Models for Authorship Verification---Notebook for PAN at CLEF 2020}}.
\newblock In \emph{{CLEF 2020 Labs and Workshops, Notebook Papers}}.
  CEUR-WS.org.

\bibitem[{Wright(2013)}]{wright2013stylistic}
David Wright. 2013.
\newblock Stylistic variation within genre conventions in the enron email
  corpus: developing a textsensitive methodology for authorship research.
\newblock \emph{International Journal of Speech, Language \& the Law}, 20(1).

\bibitem[{Wright(2017)}]{wright2017using}
David Wright. 2017.
\newblock Using word n-grams to identify authors and idiolects: A corpus
  approach to a forensic linguistic problem.
\newblock \emph{International Journal of Corpus Linguistics}, 22(2):212--241.

\bibitem[{Wright(2018)}]{wright2018idiolect}
David Wright. 2018.
\newblock \href
  {https://www.oxfordbibliographies.com/view/document/obo-9780199772810/obo-9780199772810-0225.xml}
  {Idiolect}.
\newblock \emph{Oxford Bibliographies in Linguistics}.

\end{thebibliography}
\bibliographystyle{acl_natbib}

\clearpage
\newpage
\appendix

\section{Hyperparameter tuning}\label{app:hyper}
In this section, we report the results from our hyperparameter tuning process. The following Table~\ref{tab:hyperparameters} reports some additional results obtained during hyperparameter tuning. Changing the masking probability or the margins for the contrastive loss has an impact on the final accuracy. This search is manual and not exhaustive. We used the best parameters in the main text. 

For actual implementation, we used an effective batch size of 256. The default optimizer was the Adam optimizer with a learning rate of $1e-5$. All models were trained on a single Nvidia V100 GPU with 16GB memory. The models were set to train for 5 epochs but we applied early stopping when the validation accuracy stopped to increase. Each epoch took about 2 hours to complete. For each model, we limited the maximum length of text samples to 100 tokens but the actual definition of tokens depended on the tokenizer used. 

\begin{table}[tbh]
\centering
\resizebox{0.48\textwidth}{!}{%
\begin{tabular}{lccc}
\hline
{\bf Model}  &  {\bf Accuracy} & {\bf F1} & {\bf AUC}   \\\hline
\multicolumn{4}{l}{$\tau_s=0.7$ , $\tau_d=0.3$, $\alpha=30$}\\\hline
SRoBERTa - Amazon & 0.744 & 0.676 & 0.901 \\
SRoBERTa - Reddit & 0.723 & 0.713 & 0.804\\\hline
\multicolumn{4}{l}{$\tau_s=0.8$ , $\tau_d=0.2$, $\alpha=30$}\\\hline
SRoBERTa - Amazon & 0.809 & 0.814 & 0.889 \\
SRoBERTa - Reddit & 0.708 & 0.722 & 0.784\\\hline
\multicolumn{4}{l}{$\tau_s=0.8$ , $\tau_d=0.2$, $\alpha=30$}\\\hline
SRoBERTa - Amazon & 0.809 & 0.814 & 0.889 \\
SRoBERTa - Reddit & 0.708 & 0.722 & 0.784\\\hline
\multicolumn{4}{l}{$\tau_s=0.6$ , $\tau_d=0.4$, $\alpha=10$}\\\hline
SRoBERTa - Amazon & 0.822 & 0.826 & 0.903 \\
SRoBERTa - Reddit & 0.73 & 0.734 & 0.81 \\\hline
\multicolumn{4}{l}{$\tau_s=0.6$ , $\tau_d=0.4$, $\alpha=5$}\\\hline
SRoBERTa - Amazon & 0.819 & 0.825 & 0.901 \\
SRoBERTa - Reddit & 0.72 & 0.723 & 0.798\\\hline
\end{tabular}}
\caption{Results of authorship verification on the develop / test sets}
\label{tab:hyperparameters}
\end{table}

\section{Baseline methods}\label{app:baseline}
We ran some baseline models for comparison. When setting up these models, we tried to make minimal changes to the original implementation. Details of our changes are provided below.

\myparagraph{GLAD}
We used the original code for GLAD\footnote{\url{https://github.com/pan-webis-de/huerlimann15}}. The linguistic features were extracted using the \texttt{combo4} options in the code, which covers 23 linguistic features. While support vector machine (SVM) was used as the classifier in their paper \cite{hurlimann2015glad}, we found SVM did not scale to the size of our data. Instead, we ran a logistic regression model on the features as this allowed us to interpret the feature importance. The performance of logistic regression was very close to the Random Forest classifier in their code.

\myparagraph{FVD}
The model was trained with the code released by the original authors\footnote{\url{https://github.com/pan-webis-de/weerasinghe20}} \cite{weerasinghe:2020}. We kept the original feature extraction methods and the model architecture. The input to the neural network was a 2314-dimensional feature vector, which was computed by taking the absolute difference between the linguistic feature vectors of the two authors under comparison. The two-layered fully connected neural network was trained for 100 epochs and the model with the best validation accuracy was kept.

\myparagraph{AdHomenin}
We used the original implementation\footnote{\url{https://github.com/boenninghoff/AdHominem}} provided by the author. While this implementation was slightly different from that described in \citet{boenninghoff2019explainable}, no modification was made to the code other than adapting the code to work on our own data. The same pre-processing method, model architecture, and parameters were kept. Yet the evaluation code was not used as it ignores uncertain samples, which is a standard practice in PAN 20 \cite{kestemont2020overview}. The model was trained for 5 epochs and we only kept the model with the best validation results.

\myparagraph{BERT\textsubscript{Concat}/RoBERTa\textsubscript{Concat}}
We mostly followed the description by \citet{ordonez2020will}. While \citet{ordonez2020will} used a variant of RoBERTa known as Longformer \cite{belainine-etal-2020-towards}, we re-implemented the model using the original pre-trained RoBERTa, so that the model can be directly compared to the Siamese version. Since the Longformer is highly similar to RoBERTa and BERT, we do not expect a significant performance gap between them.

\myparagraph{Evaluation}
To ensure consistency, all evaluation metrics were computed by the functions in \texttt{Sklearn}: \texttt{accuracy\_score} for accuracy, \texttt{F1\_score} for F1 and \texttt{roc\_auc\_score} for AUC.

\section{Error analysis}\label{app:error_analysis}
We also analyzed the error distributions across different conditions, shown in Table~\ref{tab:error-analysis}. Given a pair of texts, we categorized them into four different categories, same-author (SA)/different author (DA), or same domain (SD)/different domain (DD). Unsurprisingly, most methods still struggling with SA-DD and DA-SD pairs, suggesting that domain-specific/topic information partially interferes with the extraction of writing styles. The cases with RoBERTa\textsubscript{Concat} and BERT\textsubscript{Concat} are particularly interesting, as both models consistently performed worse at SA pairs but outperformed the rest of the models in DA pairs. 
Cosine distance-based models seem to better balance the trade-off across conditions. This shows that model architectures also exhibit inductive biases of their own, which may bias them to be more or less effective in certain conditions.

\begin{table}[tbh]
\centering
\resizebox{0.48\textwidth}{!}{%
\begin{tabular}{ccccc}
\hline
\multirow{2}{*}{{\bf Model}} & \multicolumn{4}{c}{{\bf Accuracy}} \\ \cline{2-5} 
 & {\bf SA-SD} & {\bf SA-DD} & {\bf DA-SD } & {\bf DA-DD }  \\ \hline
GLAD  &    73.7\%    &   58.4\%    &   62.9\%    &   73.6\%    \\ 
FVDNN & 75\% & 67.5\% & 55.7\% & 62.3\%\\
AdHominem    &  81.2\%  &   73.1\% &  63.6\% &   75.4\%    \\\hline
BERT$_{Concat}$   &  61.8\% &  51.6\% & 85.3\% & 86.5\% \\
SBERT  &  84.1\% & 77.2\%  & 67.7\% & 78\%  \\\hline
RoBERTa$_{Concat}$    &   64.4\%    &  49.6\%     &  89.2\%    & 92.5\%      \\ 
SRoBERTa & 88.4\% & 82.5\% & 74.8\% & 83.2\% \\
\hline
\end{tabular}}
\caption{Accuracy by domain and authorship. All experiments were run on the Amazon reviews.}
\label{tab:error-analysis}
\end{table}

\section{Comparing tokenization methods}\label{app:token}
\myparagraph{Tokenization methods}
For the word-based tokenization, we made use of the \texttt{word\_tokenizer} function in \texttt{NLTK}. Either 30k or 50k most frequent lexical tokens were kept as the vocabulary for training the LSTM model, plus a padding token and an OOV token. As for the BPE tokenizer, we directly used the pre-trained tokenizers for BERT and RoBERTa accessed through HuggingFace's \texttt{Transformers}.

\myparagraph{Model specification}
The underlying model is an LSTM-based Siamese network. The model consists of two bidirectional LSTM layers with 300 hidden states for each direction. The last hidden states of the last layer in both forward and backward directions were concatenated as the representation of the whole input text, which was then passed to a two-layer fully connected network with 300 hidden states in each layer. The similarity between the two paired texts was computed with the cosine distance function. The hyperparameters for the loss function were $\tau_{d}=0.4$ and $\tau_{s}=0.6$. 
No pre-trained word embedding weights were used and all weights were trained from scratch.

\myparagraph{Training details}
The training, development, and test data were the same as those in the main experiments. The model was optimized by the Adam optimizer with a learning rate of 0.001. Gradient clipping was applied to stabilize training with the maximum gradient norm set to 1. The model was trained on an 11GB RTX 2080Ti with an effective batch size of 256 for 10 epochs. The average training time for each model was about 3 hours.  

\section{Distinctiveness and consistency}\label{app:dc}
\begin{figure}
    \centering
    \includegraphics[width=\linewidth]{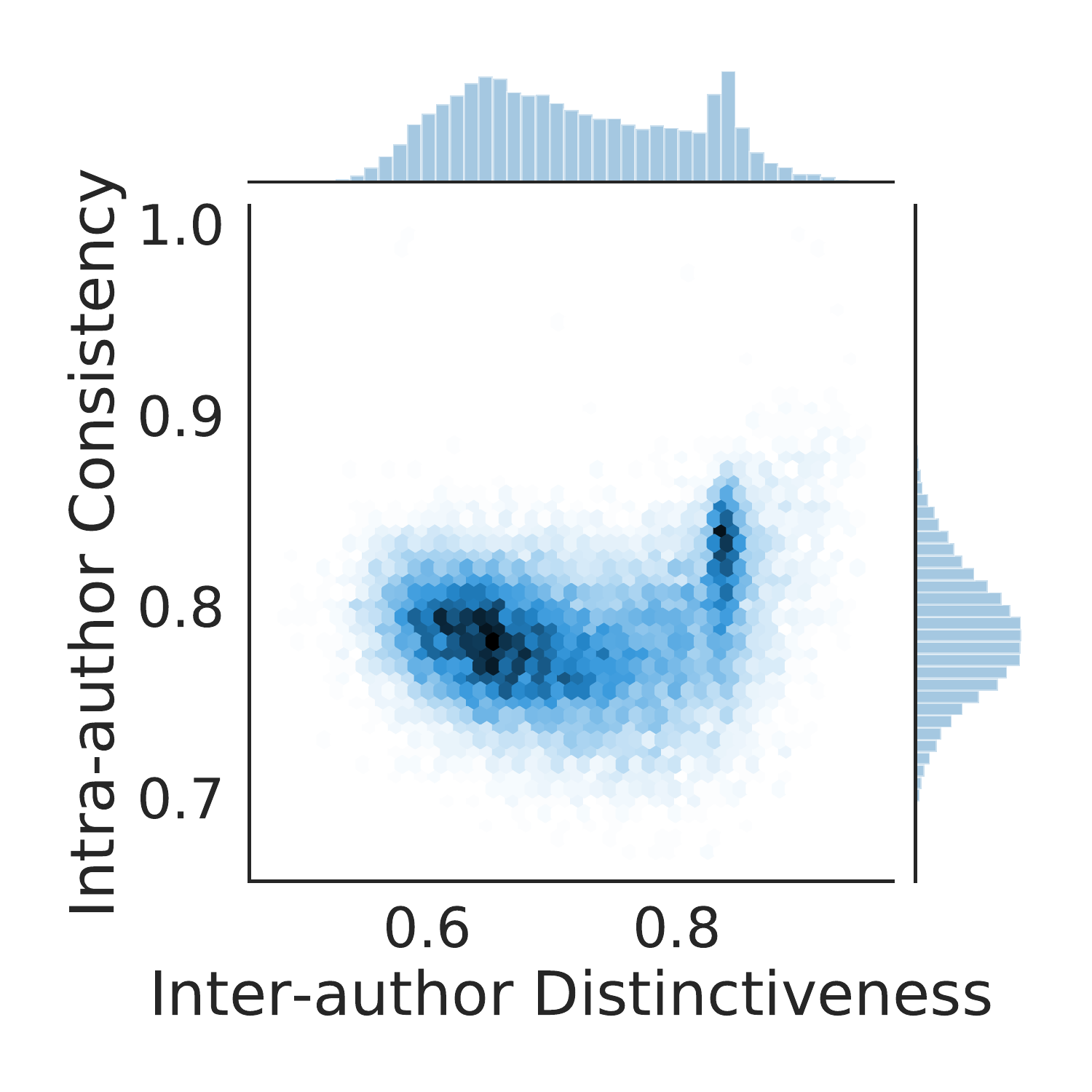}
    \caption{The joint distribution of distinctiveness and consistency as computed with Reddit posts.}
    \label{fig:bert_joint}
\end{figure}

Here we also show the joint distribution of distinctiveness and consistency given by SBERT in Figure~\ref{fig:bert_joint}. The shape of the distribution is a bivariate normal distribution and these two metrics are not correlated.  

\begin{figure}[tbh]
    \centering
    \includegraphics[width=\linewidth]{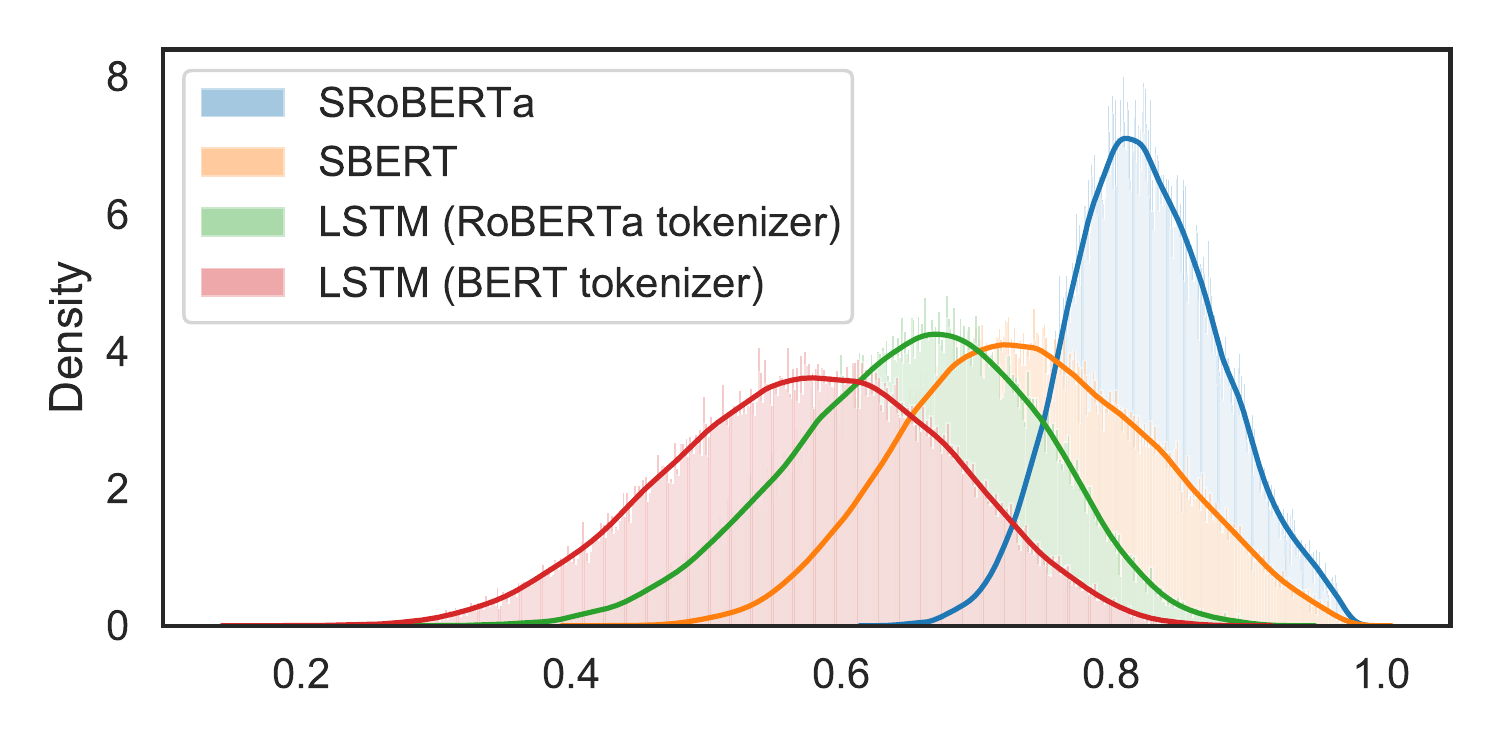}
    \includegraphics[width=\linewidth]{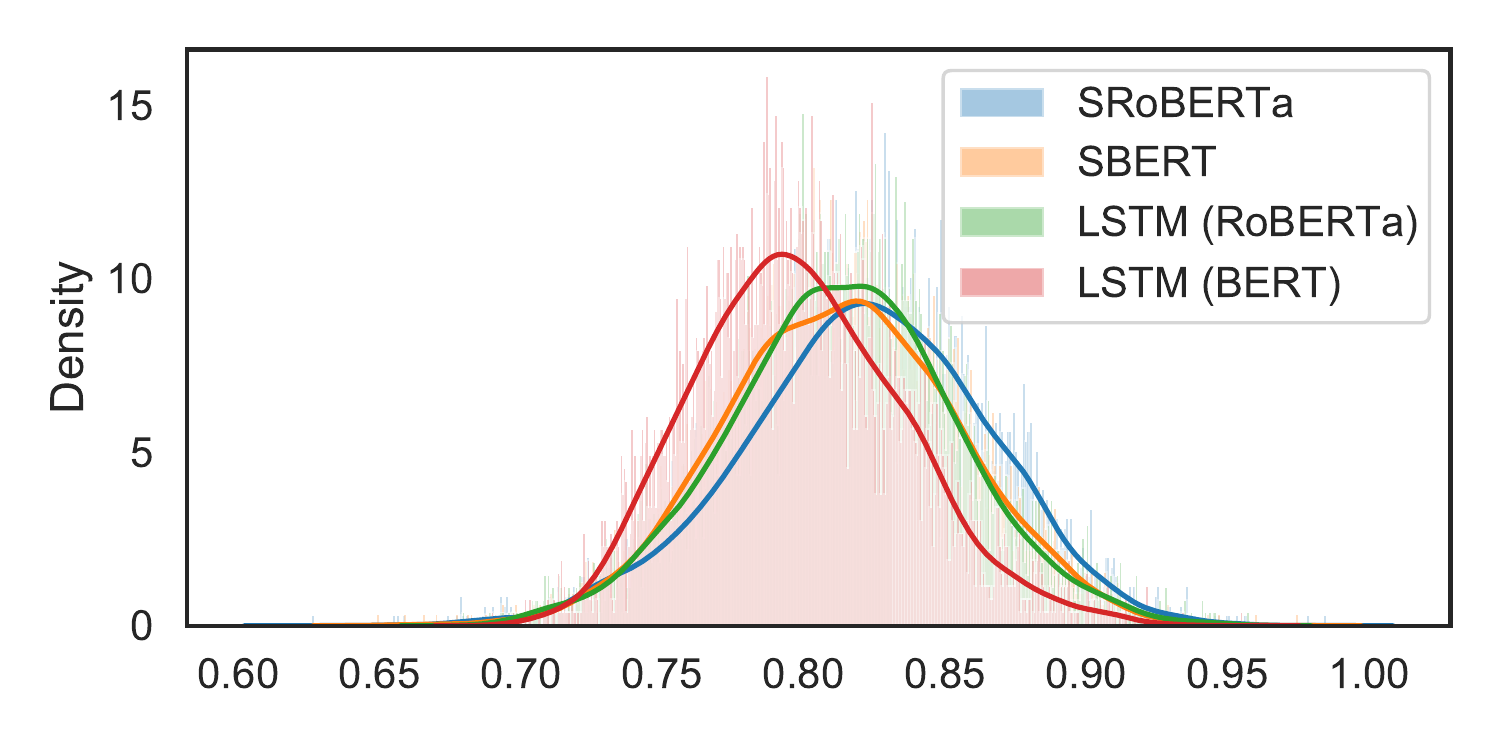}
    \caption{Distributions of style distinctiveness [{\bf Top}] and distributions of style consistency [{\bf Bottom}] in Amazon reviews.}
    \label{fig:distinct}
\end{figure}

The overall distribution of distinctiveness and consistency computed using different models are given in Figure~\ref{fig:distinct}. The distribution of style distinctiveness conforms to a normal distribution regardless of models (Figure~\ref{fig:distinct}), though the distribution is more peaked for better models. For style consistency, its distribution also conforms to a normal distribution but the distributions predicted by different models are highly similar.

\myparagraph{Correlations across models}
We used the Spearman correlation coefficients to assess to what extent different models assign similar rankings of distinctiveness and consistency. Results are presented in Table~\ref{tab:corr}.
For consistency scores given by all models are moderately correlated yet the correlations for distinctiveness are generally weak.

\begin{table}[]
\begin{tabular}{lr}\hline
Comparison & Spearman' r \\\hline
\multicolumn{2}{c}{Distinctiveness} \\\hline
SRoBERTa vs. SBERT    &    0.76                  \\
LSTM\textsubscript{RoBERTa} vs. LSTM\textsubscript{BERT}           &           0.37           \\
SRoBERTa vs. LSTM\textsubscript{RoBERTa}           &   0.12                   \\
 SBERT vs. LSTM\textsubscript{BERT}          &       0.08 \\\hline
\multicolumn{2}{c}{Consistency} \\\hline
SRoBERTa vs. SBERT    &    0.76                  \\
LSTM\textsubscript{RoBERTa} vs. LSTM\textsubscript{BERT}           &           0.61           \\
SRoBERTa vs. LSTM\textsubscript{RoBERTa}           &   0.68                   \\
 SBERT vs. LSTM\textsubscript{BERT}          &       0.58 \\\hline
\end{tabular}
\caption{Correlations for distinctiveness and consistency across model types. The results were based on Amazon reviews.}
\label{tab:corr}
\end{table}

\section{Additional analysis: characterizing sociolects}
\label{app:sociolect}

Language varies at both individual and collective levels \cite{eckert2012three}. In this section, diagnostic classification is employed to probe to what extent the collective language variations are retained in the stylistic embeddings.

\paragraph{Dataset compilation}
From the test set, we created a small subset of high socioeconomic status (SES) users and low SES users by using the prices of the reviewed products as a proxy.  We verified that there is a clear distinction in readability between high SES and low SES groups, which is a reliable linguistic indicator of SES \cite{flekova-etal-2016-exploring,basile-etal-2019-write}. 

We compiled this sociolect dataset as a subset of the test data, which contained unseen speakers and samples by the trained model. The core idea is to select users that fall into distinct socioeconomic statuses by utilizing the price tag of their reviewed products. If a user consistently reviews expensive products, it is more likely that this user is associated with high socioeconomic status. This method has been used in a study that surveyed socio-economically related variations \cite{basile-etal-2019-write}.

The meta-information was provided together with the original Amazon dataset.\footnote{\url{http://deepyeti.ucsd.edu/jianmo/amazon/index.html}} For each product, we acquired the product title and its price from the product meta-information based on its unique identifier. However, the meta-information was incomplete for a sizable fraction of data, either missing certain attributes or in the wrong format. We only kept the  products with complete meta-information. 

\begin{figure}[tbh]
    \centering
    \includegraphics[width=\linewidth]{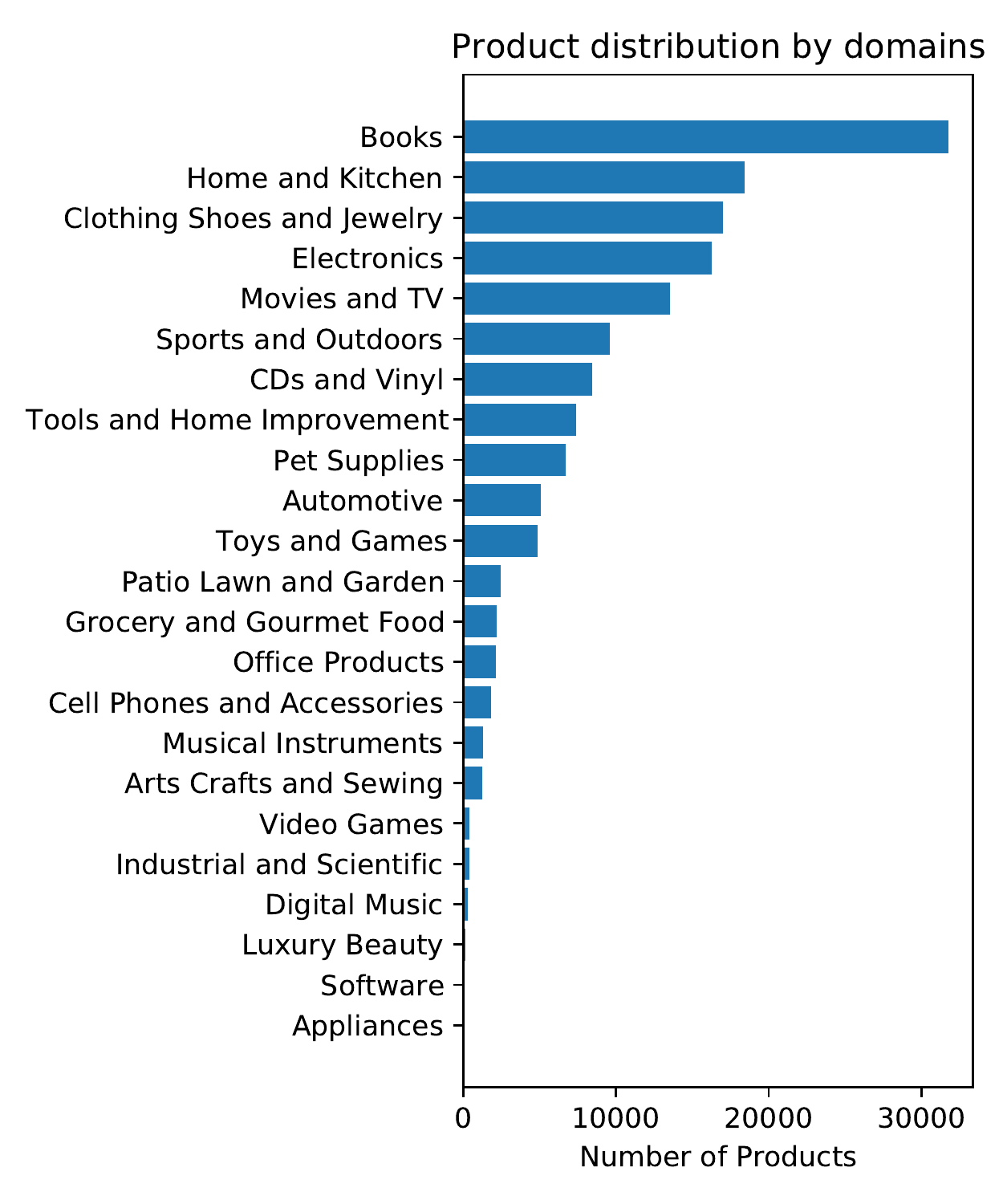}
    \caption{Distributions of product domains}
    \label{fig:product_dist}
\end{figure}

Then for each product domain, we discretized the price distribution by categorizing product prices into ten quantiles. As a proxy metric for price ranking, the quantile into which a product fell was used as an approximation of the relative expensiveness of the product. This was done for each domain separately rather than for the whole dataset, otherwise, a few domains such as appliances, luxury products, or electronics will dominate the tail of the distribution. After categorizing the data, we averaged the rankings of all the products associated with a user, the result of which was treated as an approximation of a user's socioeconomic status. We kept the top 10\% and the bottom 10\% of users as high SES users and low SES users respectively, so as to maximize the differences between these two groups. Finally, we ended up with 6567 users with 72335 reviews in the high SES group and 6939 users with 79190 reviews in the low SES groups. The dataset is relatively balanced (48\% vs. 52\%) so we did not further resample the data. The distribution of product domains is displayed in Figure~\ref{fig:product_dist}.

\paragraph{Label validation}
After dataset creation, we verified that there was linguistic variation conditioning on socio-economic status in the dataset by measuring several readability metrics, linguistic indicators shown to correlate reliably with socio-economic status \cite{flekova-etal-2016-exploring,basile-etal-2019-write}. The readability scores were computed by the functions provided in \texttt{textstat}\footnote{\url{https://github.com/shivam5992/textstat}}.
The differences are all statistically significant, implying that reviews written by high SES users tend to be more linguistically complex than those by low SES users. These results are consistent with results reported in previous studies (see Table 1 in \citet{flekova-etal-2016-exploring} and Table 3 in \citet{basile-etal-2019-write}).

\begin{table}[tbh]
\centering
\begin{tabular}{lrr}\hline
Metrics & Low SES & High SES \\\hline
ARI        &    11.8      &    13.1    \\
Coleman-Liau        &   7.26       &   7.61     \\
Dale-Chall   &   7.21       &    7.49     \\
Flesch-Reading    &    65.69      &     61.19    \\
Flesch-Kincaid        &   10.0       &   11.1      \\
Gunning-Fog & 12.04 &  13.14     \\\hline
\end{tabular}
\caption{The median values of various readability metrics.}
\end{table}

\paragraph{Training details}
Five models were trained to predict SES based on language: 1) TF-IDF, with mostly topical information; 2) Handcrafted stylometric features \cite{weerasinghe:2020}; 3) RoBERTa with both topical and linguistic features; 4) SRoBERTa embeddings with idiolectal features and 5) a random baseline (BL). We also used the same models and data to predict the product domain of each short text. 

For TD-IDF and Stylometric features, we used logistic regression as the base model. The stylometric features were extracted using the FVD method \cite{weerasinghe:2020}, one of our baseline methods for authorship verification. For RoBERTa and SRoBERTa, we added a two-layered neural network on top of the \texttt{[cls]} token with cross-entropy loss. The only difference was that, for SRoBERTa, the base RoBERTa was freezed during training. 
We ran each model 3 times with different random seeds. For each time, we randomly split the data into 75\% and 25\% partitions for training and testing. The averaged results were reported.

\begin{table}[tbh]
\resizebox{0.48\textwidth}{!}{%
\begin{tabular}{lrrrrr}\hline
Model  & TF-IDF & Stylo. & RoBERTa & SRoBERTa & BL\\\hline
SES    &   0.633    &  0.588   &  \textbf{0.644}    &   0.592 & 0.50\\
Domain &   0.601     &   0.492  &   \textbf{0.681}       & 0.343  & 0.03     \\\hline
\end{tabular}}
\caption{F1  scores for SES and domain predictions. }
\label{tab:group}
\end{table}

\paragraph{Results} 
For the challenging task of SES prediction, all models attain moderate performance that is consistently above chance level (Table~\ref{tab:group}), echoing previous findings \cite{flekova-etal-2016-exploring,basile-etal-2019-write}. Compared to the fine-tuned RoBERTa, the idiolectal features have filtered out some SES-related variations, which could be related to domain-specific information.
Notably, the style embeddings performed the worst at predicting product domain indicates the idiolectal style is not simply capturing product domain as a proxy for SES (e.g., learning more expensive domains). The SRoBERTa's high performance on SES and low performance on domain suggest that our task setup and sampling strategy forced the model to smooth out a significant portion of variation associated with topics. As noted by \citet{boenninghoff2019explainable}, even if surface linguistic features are not highly content-related, they still achieve  moderate performance, suggesting that variation across domains may be more than topical. 
The fact that SES variations are present in the idiolectal embeddings suggests that at least some SES variations are nested within idiolectal variation \cite{eckert2012three}.

\section{Additional text samples}
Additional samples pf Amazon reviews with polarizing distinctiveness are given in Table~\ref{app:samples}. Different models single out reviews with wide-ranging stylistic traits. Full reviews are shown in the table, though only the first 100 words are used by the model during inference. 

Table~\ref{tab:consistent} shows the text samples from the most and the least consistent authors in terms of their writing styles. For each model, each column presents reviews written by the same author.

\begin{table*}[tbh]
\begin{tabular}{lp{6.5cm}p{6.5cm}}
{\bf Model} & {\bf Most distinctive } & {\bf Least distinctive}  \\\hline
LSTM\textsubscript{BERT} & {\small The late \'80s were a golden age for CD reissues, especially of tracks from the \'50s and \'60s, since the new digital format was just gaining popularity, there was a retro-1960s revival going on, and record companies realized they had whole new revenue stream from people buying (or re-buying) back-catalog material for their new players. The compilations issued then were full of quality stuff, unlike later bottom-of-the-barrel reissues. } & {\small This Urban Fantasy series pulls you right in and the more you know the characters the more you want to know. Hailey Edwards will make you smirk, bite your nails, cry and hope, hope, hope because her characters become (our) friends. As fantastic as the characters origins and abilities are their personalities are so appealing that I found myself hoping in the goodness of even some of the meanies. } \\\hline
LSTM\textsubscript{RoBERTa} & {\small **UPDATE 4/19/16** apparently got a bad cable Couldn't figure out why I was having issues connected to Ethernet Thought it might be a network driver issue or a modem issue But after replacing this cable with a shorter one had laying around come to conclusion its this cable that was bad Not a big deal it happens only out a couple \$\$\$ , disappointed but not to upset  I need a 15 foot+ Ethernet cable it works , really not much to review ends snap in ok , no twists in cable works good} & {\small My cats don't like to be brushed. But when I can get several strokes in, this works well. I use the dog brush on my dog; the cat brush is a little smaller that the dog brush and weighs less which are good changes to make for the kitty models.} \\\hline
SBERT    & {\small A Great Forza,  Serafin conducts with wonderful pace, warmth and subtly for such an unsubtle opera making this a real beauty, and so easy to listen to. Callas is quite magnificent with a fine supporting cast. Disregard many of the somewhat breathless negatives, gushing with crushes and arguments for other favorite sopranos, so juvenile, the fact is there are many great female opera singers all suited to different operas some more than others, Callas happens to be one of the greatest in emotional commitment and inner depth of feeling, } 
& {\small So I hold a bachelors and masters in Speech Language Pathology and really have limited background in computer programing. Even with my good command of the English Language, I found this book difficult to follow and found myself rereading sections of it.  I had to get through a third of the book, just to have an idea of what it was about. I learned about the history of computer programing and the need for there to be a better system for programmers and managers to communicate and produce better outcomes. } \\\hline
SRoBERTa &   {\small WHAT a WASTE of TIME !!!  The LARGEST Funnel  ... Is " MAYBE  3 Inch WIDE " \&  the TUBE Part,  MIGHT  can  FIT a  \# 2 PENCIL in IT ??? The SMALL  ONE  has  a  TUBE  With LESS Then 1/4 Inch ???  AS  a  COOK ,  THESE  are " A TOTAL JOKE " ( SOMETHING , I SEE at a FLEA MARKET) !!!  WHAT..... "FOODS  CANYOU FIT 1/4 - 1/3 INCH OPENING " ???}   & {\small I wanted a simple steel men's ring without a design and that wouldn't show fingerprints. This ring is perfect. One great thing that I enjoy is that the interior is rounded and polished, making it feel like silk when I put it on. Very affordable, too! Just goes to show, you don't have to break the bank to get attractive quality.} \\\hline                                                     
\end{tabular}
\caption{Sample review excerpts with the most and the least distinctive style.}
\label{app:samples}
\end{table*}

\begin{table*}[]
\centering
\begin{tabular}{lp{6.5cm}p{6.5cm}}
{\bf Model} & {\bf Most consistent} & {\bf Least consistent} \\\hline
\multirow{2}{*}{BERT}  & {\small {\bf Author A:} capcom is the greatest video game company in the universe there true genius's the best of the best capcom rocks all the games capcom made from the 80s,90s,2000s,2010's and 2015 are the greatest video games in the universe there true classics the best of the best all the games capcom made from the 80s,90s,2000s,2010;s and 2015 rocks 2015 is the greatest year for capcom a perfect year the best of the best 2015 for capcom rocks \textasciicircum\_\textasciicircum }        & {\small  {\bf Author B:} I haven't read the novel. I can't say whether this is a good adaptation. There is no plot as such; just a random collection of events dictated by fate. When it became clear that all characters are pawns of fate, what happens to them became uninteresting. Likely, this follows the novels intention. I watched for 30 min and stopped.}                 \\\cline{2-3}
       & {\small  {\bf Author A:} the star wars prequel trilogy is the greatest movie trilogy in the universe there true classics the best of the best the star wars prequel trilogy rocks the star wars charecter anakin skywalker is the greatest movie charecter in the universe its pure genius the best of the best the star wars charecter anakin skywalker rocks \textasciicircum\_\textasciicircum }        &     {\small  {\bf Author B:} A true feat of alchemy, turning base metal (a script worth it\'s weight in manure) into piles of cash. Or more specifically, this is one of the dumbest, least plausible, movies we\'ve watched in a long time. And yet not without comic relief. Now, who was it that said, "Nothing will come of nothing"? Silly old bard.}     \\\hline
\multirow{2}{*}{RoBERTa}  &   {\small {\bf  Author C:} these our not real instruction tapes but introductions to who Larry really is you can learn from them some really good stuff kenpo is a marshal art that is based on common sense any one who really understand his marshal art will be doing kenpo with out knowing kenpo our even taking a class all marshal artist will run it to these principles for they our the principle of the sword}      &   {\small  {\bf  Author D:} Some of these songs have only appeared on CBS Christmas compliation albums in the 60\'s.  I have not run across the Mike Douglas "Touch Hands on Christmas Morning" since it\'s appearance on a CSP LP for Grant\'s Department Stores in 1967.  Audio quality is pretty decent for the age of the recordings.  The shear quantity of the music makes it well worth \$14.  Happy Holidays.}           \\\cline{2-3}
       & {\small  {\bf  Author C:} i set down with the book try to read it and this is a book that takes a lot of time to read but in it i find nothing really new our enlightening just same old new age dribble it sounds like a psychologist is writing the book i find the same old thing in new age books now there is a lot of people that would say this is a good book}  & {\small {\bf  Author D:} Well worth the price.  Helps greatly compared to a junky rubber duckie antenna that comes with portables.  Got mine with the SMA connector which works on Yaesu portables.  You will need an SMA female to female SMA adapter to use with the cheap Chinese portables. Even works on mobiles if you keep the power down and use an adapter.  A good choice for scanner use on VHF \& UHF too}    \\\hline       
\end{tabular}
\caption{The most and the least consistent authors as identified by SRoBERTa and SBERT.}
\label{tab:consistent}
\end{table*}

\end{document}